\documentclass[twocolumn]{m3ailpreprint}
\usepackage{colortbl}
\usepackage[shortlabels,inline]{enumitem}
\usepackage[linesnumbered,ruled,vlined]{algorithm2e}

\colorlet{cvprblue}{mthreeblue}
\definecolor{tbhead}{HTML}{F3F9FE}
\definecolor{tbzebra}{HTML}{F9FCFF}

\title{A Simple Yet Effective Agent for Open-World Aerial Object Search}

\author[1,\ast]{Jitai Hao}
\author[1,\ast]{Liangxing Yang}
\author[2]{Zhenyang Li}
\author[3]{Fan Liu}
\author[1,\dagger]{Qiang Huang}
\author[1,\dagger]{Jun Yu}

\affiliation[1]{Harbin Institute of Technology (hit.edu.cn)}
\affiliation[2]{Hong Kong Polytechnic University (polyu.edu.hk)}
\affiliation[3]{\mbox{Southeast University (seu.edu.cn)}}
\contribution{$^\ast$Equal contribution.}
\contribution{$^\dagger$Corresponding authors.}
\abstract{
Open-world aerial object search methods often rely on persistent spatial maps, learned world models, or task-specific policy training.
To examine how far a minimal, training-free design can go, we introduce \textbf{SimpleUAV}, an agent built around a frozen, off-the-shelf vision-language model (VLM). 
SimpleUAV separates current-view target assessment from history-conditioned action selection, combining fresh candidate evidence with lightweight search context.
Local geometry converts visual selections into bounded straight-line movements, depth-based checks halt blocked segments, and measured execution feedback guides subsequent decisions. 
This closed loop sustains search without persistent spatial maps or task-specific training.
SimpleUAV achieves \textbf{41.61\%} Success Rate (SR) on UAV-ON and \textbf{11.98\%} on AerialDojo-200K, compared with \textbf{23.80\%} and \textbf{3.20\%} for the strongest prior baselines, respectively. 
With limited benchmark-specific adaptation, the same core agent transfers to aerial Vision-and-Language Navigation (VLN) on TravelUAV, improving Full SR by at least \textbf{19.66\%} across all three splits under L1 assistance.
These results establish a strong baseline for evaluating when additional architectural complexity benefits aerial search.
}

\checkdata[Code]{\url{https://github.com/beilu-zhuma/SimpleUAV}}
\correspondence{\email{huangqiang@hit.edu.cn}, \email{yujun@hit.edu.cn}}

\begin{document}

\onecolumn
\enlargethispage{1.25em}
\mymaketitle[{%
  \noindent\begin{minipage}{\textwidth}
    \centering
    \captionsetup{hypcap=false}
    \includegraphics[width=0.99\textwidth]{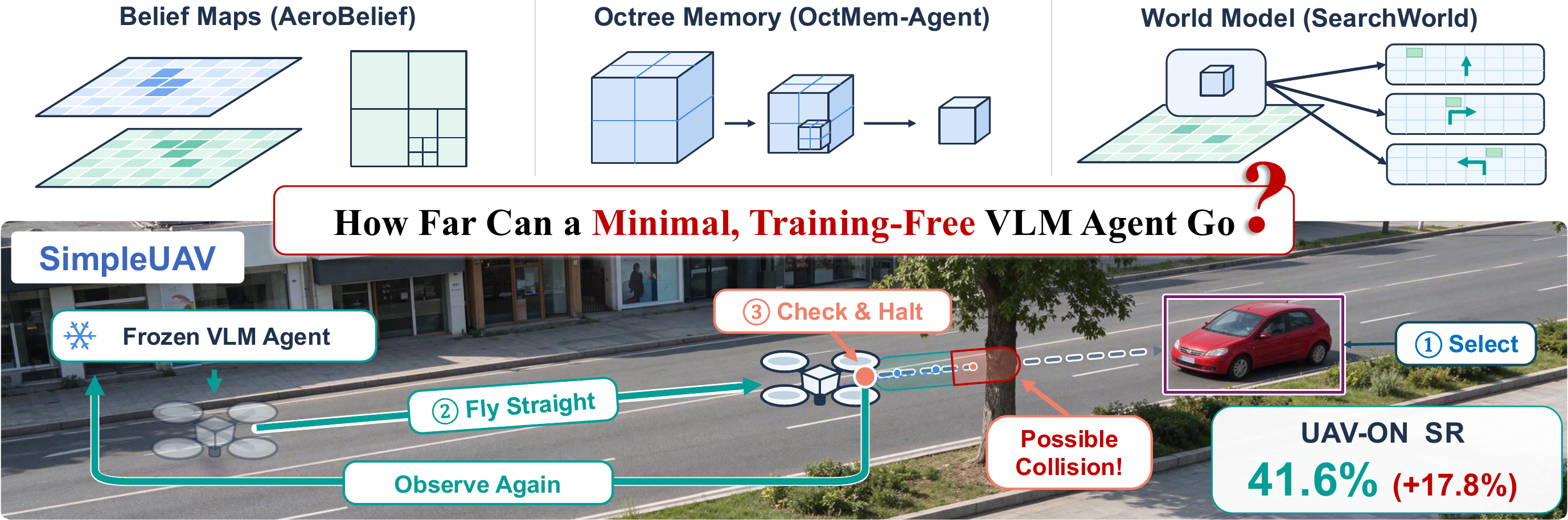}
    \captionof{figure}{\textbf{Stronger aerial search with a minimal agent.} Representative prior systems (top) use semantic belief maps, explicit spatial memory, or learned world models.
    \textbf{SimpleUAV} (bottom) uses a frozen VLM in a local depth-guided closed loop, requiring neither persistent spatial maps nor task-specific training, yet improves success rate (SR) by \textbf{17.8\%} over the strongest prior result.}
    \label{fig:teaser}
  \end{minipage}\par
  \vspace{0.65em}
}]

\clearpage
\begingroup
\setlength{\textwidth}{6.875in}
\setlength{\columnsep}{0.3125in}
\setlength{\oddsidemargin}{\dimexpr(\paperwidth-\textwidth)/2-1in-\hoffset\relax}
\setlength{\evensidemargin}{\oddsidemargin}
\renewcommand{\rmdefault}{ptm}
\normalfont
\twocolumn
\raggedbottom
\section{Introduction}
\label{sec:intro}

Aerial object-goal navigation requires an unmanned aerial vehicle (UAV) to locate and approach a target described in natural language in an unfamiliar outdoor environment~\cite{xiao2025uavon, ji2026cityavos, feng2026aerialdojo}.
This capability supports applications such as search and rescue~\cite{wu2023uavrescue} and infrastructure inspection~\cite{tse2024uavinspection}.
Given neither the target location nor detailed route instructions \cite{zhou2026octmem}, the UAV must decide where to search, which observations warrant further inspection, and when the available evidence supports stopping \cite{ji2026searchworld}.

These decisions are tightly coupled in large, partially observed outdoor environments.
The target may remain outside the UAV's field of view for much of an episode, requiring contextual cues to guide exploration~\cite{kim2026raven,xiao2026aerobelief}.
Even if a candidate appears, distance, occlusion, and visually similar distractors can make its identity uncertain~\cite{yang2026skyagent}.
Moving closer or changing altitude may reveal discriminative attributes, but also changes the visible scene and exposes the UAV to new obstacles~\cite{chen2026airhunt}.
Thus, effective search requires semantic judgments, search context, and three-dimensional motion to remain coordinated as observations evolve.

Recent aerial search systems address these challenges with specialized spatial representations and predictive models. 
Three representative directions are shown in Figure~\ref{fig:teaser}:
\begin{itemize}[leftmargin=*,itemsep=2pt,topsep=3pt]
  \item \textbf{Semantic Maps and Beliefs.} 
  Task-oriented semantic, attraction, uncertainty, or value maps turn local observations into sustained search guidance ~\cite{ji2026cityavos, zhang2026apex, chen2026airhunt}. 
  AeroBelief further separates contextual cues from target-specific evidence through dual-layer semantic-spatial belief maps~\cite{xiao2026aerobelief}.
  
  \item \textbf{Explicit Spatial Memory.} 
  Semantic voxel-ray memory \cite{kim2026raven}, sliding local maps with sparse global history \cite{hou2026slider}, and adaptive octree memory with instruction-guided queries \cite{zhou2026octmem} organize fragmented observations across changing viewpoints.
  
  \item \textbf{Learned World Models.} 
  SearchWorld combines explicit spatial memory with learned dynamics to reason about action consequences and improve its exploration policy through imagined rollouts~\cite{ji2026searchworld}.
\end{itemize}

These approaches introduce persistent representations, dedicated memory-processing modules, or task-specific model and policy training.
At the same time, pretrained vision-language models (VLMs) offer broad semantic knowledge and open-vocabulary visual understanding~\cite{bai2025qwen3, qwen36_35b_a3b}.
These capabilities provide a basis for interpreting target descriptions, identifying promising search regions, and assessing visible candidates without navigation-specific fine-tuning.
They motivate a systematic evaluation of how effectively a frozen VLM can support aerial search when paired with lightweight search context and local geometric execution.
This leads us to ask:

\par
\vspace{0.5em}
\begingroup
\setlength{\fboxsep}{0em}%
\noindent\colorbox{cvprblue!6}{%
  {\color{cvprblue!80}\vrule width 2.0pt}%
  \hspace{0.8em}%
  \begin{minipage}{\dimexpr\linewidth-4pt-1.6em\relax}
    \vspace{0.6em}
    \itshape
    How far can a minimal, training-free VLM-based agent go in open-world aerial search?
    \par\vspace{0.6em}
  \end{minipage}%
  \hspace{0.8em}%
}\par
\vspace{0.5em}
\endgroup

We investigate this question with \textbf{SimpleUAV}, a closed-loop agent built around a frozen, off-the-shelf VLM.
The VLM first assesses candidate targets from current multi-view observations alone, then selects actions using localized candidates, recent interactions, and a bounded text summary. 
This separates current visual assessment from the context needed to maintain search continuity. 
Local geometry converts visual selections into bounded straight-line movements, while depth-based obstacle checks halt blocked segments. 
Measured execution outcomes inform the next observation and decision cycle, connecting semantic reasoning to actual motion without persistent spatial maps or task-specific policy training.

Without task-specific fine-tuning, SimpleUAV improves success rate (SR) over the strongest compared baselines by \textbf{17.81\%} on UAV-ON \cite{xiao2025uavon} and \textbf{8.78\%} on AerialDojo-200K \cite{feng2026aerialdojo}. 
On the aerial VLN benchmark TravelUAV \cite{wang2024openuav}, it improves Full SR by at least \textbf{19.66\%} across all three evaluation splits under L1 assistance.
Even with a compact 4B VLM, SimpleUAV exceeds the strongest prior UAV-ON baseline by \textbf{14.13\%} in SR.

Our main contributions are summarized below.
\begin{itemize}[leftmargin=*,itemsep=2pt,topsep=3pt]
  \item We show that a \textbf{minimal, training-free} agent built around a frozen VLM can substantially outperform specialized systems in open-world aerial search without persistent spatial maps or task-specific policy training.
  
  \item We propose \textbf{SimpleUAV}, separating current-view target assessment from history-conditioned action selection and closing the loop through local geometric execution feedback to sustain search, while requiring fresh visual and geometric evidence for approach and stopping.
  
  \item We demonstrate \textbf{generalization} across environments and navigation tasks, transferring SimpleUAV from UAV-ON to AerialDojo-200K and TravelUAV with limited benchmark-specific adaptation and no additional training, achieving the highest SR among compared methods.
\end{itemize}

\section{Related Work}
\label{sec:related_work}

\paragraph{Aerial Object-Goal Navigation.}
Indoor ObjectNav has expanded from category-conditioned exploration~\cite{chaplot2020semexp, ramakrishnan2022poni} to zero-shot open-vocabulary search~\cite{gadre2023cow}, distinguishing unseen-object search from navigation to visible candidates~\cite{dang2023dual}.
Aerial benchmarks specify instance-level language or image goals \cite{xiao2025uavon, ji2026cityavos, feng2026aerialdojo}. 
Exploration uses frontiers \cite{yamauchi1997frontier} or viewpoint planning based on target visibility and search history (Star-Searcher \cite{luo2024starsearcher}). 
AOA \cite{xiao2025uavon} prompts movement selection from onboard observations and recent poses.
Aerial systems aggregate cross-view evidence using geometric memory (octrees, voxel grids, and sliding local maps) \cite{zhou2026octmem, kim2026raven, hou2026slider} or semantic maps that prioritize regions and retain target evidence \cite{zhang2026apex, ji2026cityavos, chen2026airhunt, xiao2026aerobelief}.
AeroBelief \cite{xiao2026aerobelief} separates contextual cues from target beliefs, while SkyAgent \cite{yang2026skyagent} revises hypotheses under conflicting observations. Predictive methods estimate unseen semantics, action outcomes, or future aerial views \cite{zhang2024imagine, nie2025wmnav, zhang2025anwm}, with SearchWorld \cite{ji2026searchworld} learning its exploration policy from imagined trajectories.
SimpleUAV sustains search with a frozen VLM, local depth geometry, recent observations, execution feedback, and a bounded text summary, without persistent spatial maps, learned world models, or policy training.

\paragraph{VLM-based Vision-and-Language Navigation (VLN).}
VLN maps route instructions to sequential actions, from indoor viewpoint graphs (Room-to-Room~\cite{anderson2018r2r}) to continuous control (VLN-CE~\cite{krantz2020vlnce}).
Advances include richer instruction supervision and longer observation histories~\cite{fried2018speaker, ku2020rxr, chen2021hamt}, prompting with textual history, topological maps, or specialized experts \cite{zhou2024navgpt, chen2024mapgpt, long2024discussnav}, and trained action prediction from pretrained multimodal representations \cite{zhou2024navgpt2, zhang2024navid}.
Aerial VLN supports simulated and real 3D flight~\cite{liu2023aerialvln, wang2024openuav, lee2025citynav} through spatial representations linking landmarks to motion~\cite{gao2025stmr, zhang2025citynavagent, jiang2026spatialfly, xu2026geonav}, learned multimodal policies \cite{cai2025flightgpt, gao2025openfly, zhang2025navfom}, or observation histories and predicted views or spatial maps \cite{wang2023dreamwalker, jiang2025longfly, wang2026forefly, liu2026airforesight}.
SkyVLN~\cite{li2025skyvln} couples language planning with nonlinear model predictive control.
SimpleUAV addresses open-world object search by assessing candidates solely from current views and selecting actions using recent search context. Approaching requires a visible, depth-localized candidate, while stopping further requires a confirmed match within stopping range.

\paragraph{Language-Guided Embodied Agents.}
Language-guided agents ground semantic decisions in executable actions and revise plans through feedback. 
SayCan \cite{ichter2023saycan} assesses skill feasibility through learned affordances, while \mbox{LM-Nav}~\cite{shah2023lmnav} uses a pretrained policy to navigate between visually grounded landmarks. 
Reflexion \cite{shinn2023reflexion} uses textual feedback across attempts, while Inner Monologue \cite{huang2023innermonologue} feeds execution outcomes and scene observations to the planner within tasks.
Object-search agents update search state through language-model reasoning, refine localization through new viewpoints, or replan when observations contradict semantic-map predictions \cite{cao2025cognav, qin2025aprr, ko2025care}. 
PSC-AVDN \cite{qi2026pscavdn} shares spatial memory between separate aerial exploration and confirmation stages.
SimpleUAV guides VLM search after incomplete or blocked movements using measured displacement and segment termination status from bounded straight-line motion with local depth checks.

\section{Method}
\label{sec:method}

\begin{figure*}[t]
\centering 
\includegraphics[width=\textwidth]{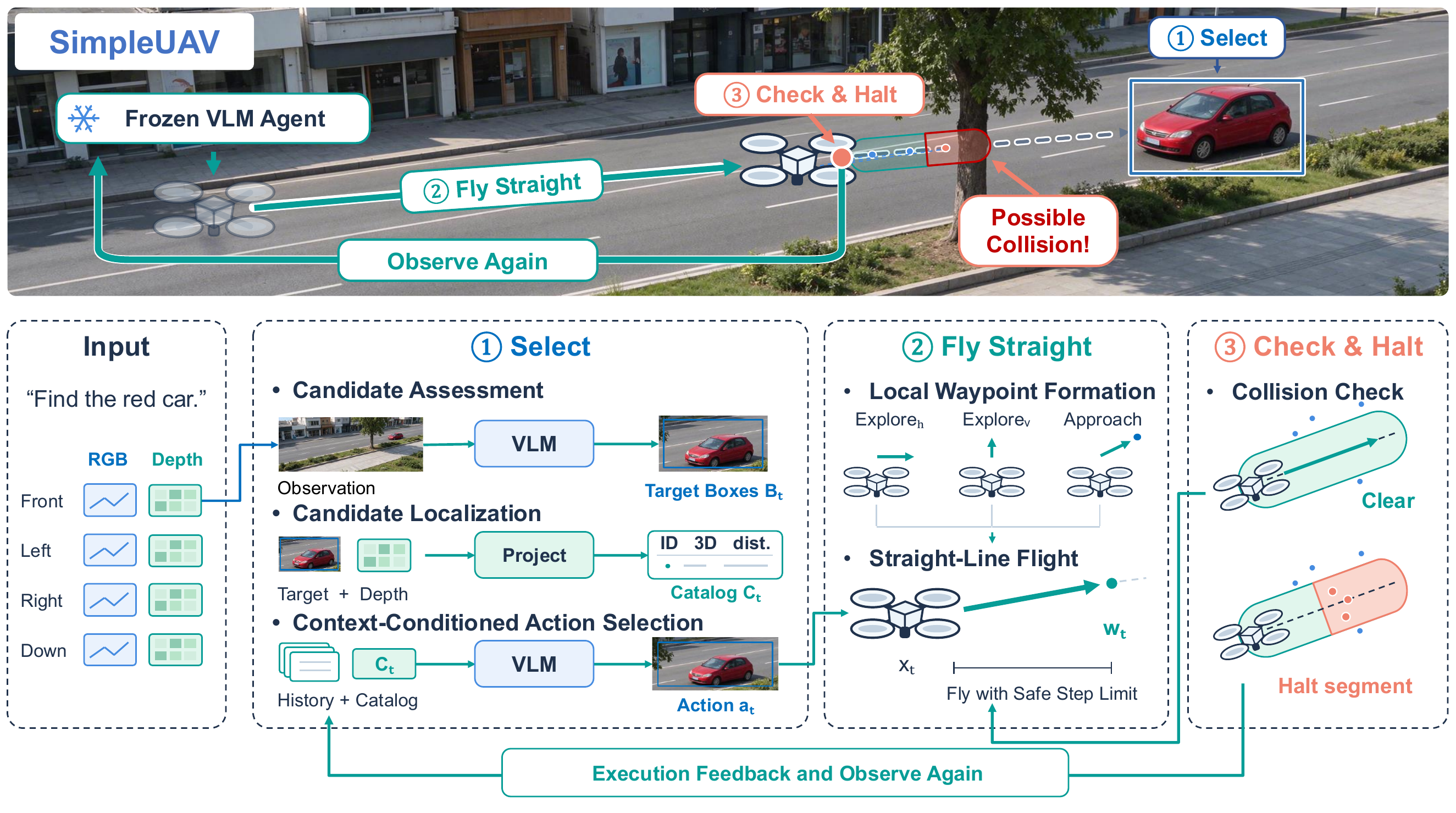}
\vspace{-2.0em}
\caption{\textbf{Architecture of SimpleUAV.}
Given a language goal and four RGB-D views (front, left, right, and downward), SimpleUAV executes three stages without persistent maps or task-specific training. 
\textbf{(1) Select (Section~\ref{sec:method_selection}):} A frozen VLM detects candidate boxes ($\mathcal{B}_t$) from current RGB views. 
Depth and camera calibration provide 3D candidate locations to construct ($\mathcal{C}_t$), and a second VLM query selects an action ($a_t$) using this catalog and lightweight interaction history. 
\textbf{(2) Fly Straight (Section~\ref{sec:method_flight}):} The selected region or candidate determines a 3D waypoint ($w_t$) for exploration ($\text{Explore}_h$, $\text{Explore}_v$) or $\text{Approach}$, and the UAV moves from $\bm{x}_t$ along a bounded straight-line segment.
\textbf{(3) Check \& Halt (Section~\ref{sec:method_safety}):} Throughout motion, local depth checks test upcoming segments using a 3D safety capsule and halt the current segment when an obstacle is detected.
Execution feedback guides the next observation and decision cycle.}
\label{fig:method}
\vspace{-0.5em}
\end{figure*}

Aerial search requires the UAV to assess visible candidates, retain useful search context, and respond to obstacles during motion.
To coordinate these requirements, the \textbf{SimpleUAV} agent couples a frozen VLM with local depth geometry in a recurrent Select-Fly Straight-Check and Halt loop, as shown in Figure \ref{fig:method}.

\subsection{Problem Formulation}
\label{sec:method_task}

Given a language goal $g$ describing a target object, the UAV starts at position $\bm{x}_0$ and must explore, approach the target, and stop within a decision budget.
At decision step $t$, it receives its position $\bm{x}_t$ and four RGB-D views with camera poses, giving the observation $\mathcal{O}_t$:
\begin{equation}
  \mathcal{O}_t = \bigl(
    \bm{x}_t, \{(I_t^v, D_t^v, \bm{T}_{wc,t}^v)\}_{v \in \mathcal{V}}
  \bigr).
  \label{eq:observation}
\end{equation}
Here, the position $\bm{x}_t$ lies in $\mathbb{R}^3$, and $\mathcal{V}$ indexes the \texttt{front}, \texttt{left}, \texttt{right}, and \texttt{downward} cameras.
For view $v$, $I_t^v$ and $D_t^v$ are the RGB and depth images, and $\bm{T}_{wc,t}^v$ is the homogeneous camera-to-world transform.

The VLM receives RGB images and textual context, while the program uses depth and calibration to locate candidates and check motion.
Success requires a collision-free episode ending with an accepted stop within the target's evaluation radius. 
The ground-truth target position is used only for evaluation.
The high-level action $a_t$ has a type in:
\begin{equation}
  \mathcal{A} 
  = 
  \{\text{\texttt{Explore}}_h, \text{\texttt{Explore}}_v, \text{\texttt{Approach}}, \text{\texttt{Stop}}\}.
  \label{eq:action_types}
\end{equation}
Each action specifies a view and image region.
Vertical exploration additionally specifies ascent or descent, while \texttt{Approach} and \texttt{Stop} reference a candidate identifier in the current candidate catalog.
Executing each movement action traces a single step-bounded straight-line segment via the controller's low-level velocity commands.

\subsection{Select}
\label{sec:method_selection}

At each decision step $t$, the UAV selects an intended region to inspect or approach, without maintaining persistent spatial maps.
Rather than training a separate object detector or conflating target verification with search planning, we use two sequential queries to the same frozen VLM $F_\theta$: the first performs stateless candidate assessment from the current views, and the second selects an action using the localized candidates and recent search history.

\paragraph{Candidate Assessment.}
Let $\mathcal{I}_t=\{I_t^v\}_{v\in\mathcal{V}}$ denote the current RGB views.
Under the target assessment prompt $\pi_{\mathrm{check}}$, the VLM produces:
\begin{equation}
  \mathcal{B}_t 
  = 
  F_\theta(g,\mathcal{I}_t;\pi_{\mathrm{check}}).
  \label{eq:target_assessment}
\end{equation}
This query excludes historical images, search memory, and previous decisions to keep current candidate assessment separate from earlier search hypotheses.
Each entry records the candidate’s view, bounding box, visible attributes, and match label. 
A possible candidate needs further inspection, while a confirmed candidate has visible attributes judged by the VLM to match $g$.
The list may be empty. Candidate identifiers are assigned anew at each step, with no programmatic identity tracking across decisions.

\paragraph{Candidate Localization.}
Image regions identify potential targets, but approach and stopping decisions also require geometric estimates of their locations and distances.
We therefore localize each candidate in $\mathcal{B}_t$ using onboard depth and camera calibration, using the resulting geometry locally.

For candidate $i$ in view $v_i$, we select a pixel $\bm{u}_{t,i}$ corresponding to the 25th percentile of valid radial depth samples within its bounding box in $D_t^{v_i}$, and denote its depth by $d_{t,i}$.
Scaling the calibrated unit camera ray $\hat{\bm{r}}_{v_i}(\bm{u}_{t,i})$ by $d_{t,i}$ and applying the camera pose gives:
\begin{equation}
    \begin{aligned}
        \begin{bmatrix} \hat{\bm{p}}_{t,i}^{w} \\ 1 \end{bmatrix}
        &= \bm{T}_{wc,t}^{v_i}
        \begin{bmatrix}d_{t,i}\hat{\bm{r}}_{v_i}(\bm{u}_{t,i}) \\ 1\end{bmatrix}, \\
        \hat d_{t,i} &= \|\hat{\bm{p}}_{t,i}^{w}-\bm{x}_t\|_2.
    \end{aligned}
    \label{eq:candidate_grounding}
\end{equation}
Here, $\hat{\bm{p}}_{t,i}^{w}$ is the estimated 3D surface point in world coordinates, and $\hat{d}_{t,i}$ is its distance from the UAV.
Candidates without valid depth remain available to the action prompt as visual cues for exploration but are ineligible for \texttt{Approach} or \texttt{Stop}. 
The resulting catalog $\mathcal{C}_t$ records localization validity and estimated distance for action selection.

\paragraph{Context-Conditioned Action Selection.}
Selecting the next action requires both current candidates and a compact record of prior search progress. 
Thus, the agent constructs the navigation context:
\begin{equation}
  \mathcal{K}_t = (\bm{x}_t, \mathcal{C}_t, \mathcal{H}_t, m_{t-1}, f_{t-1}),
  \label{eq:navigation_context_Kt}
\end{equation}
where $\mathcal{C}_t$ is the catalog of currently detected candidates; $\mathcal{H}_t$ contains recent selected-view images and their executed actions; and $f_{t-1}$ provides the latest execution outcome.
The bounded text summary $m_{t-1}$ maintains search continuity by tracking inspected regions, unresolved hypotheses, and failed directions.

Under the navigation prompt $\pi_{\mathrm{nav}}$, the VLM selects the high-level action $a_t$ and updates the summary to $m_t$:
\begin{equation}
    (a_t,m_t) = F_\theta(g,\mathcal{I}_t,\mathcal{K}_t;\pi_{\mathrm{nav}}).
    \label{eq:context_navigation}
\end{equation}
For exploration actions ($\text{\texttt{Explore}}_h$ or $\text{\texttt{Explore}}_v$), the VLM designates an uninspected direction or image region.
For \text{\texttt{Approach}} and \text{\texttt{Stop}}, the output must reference a currently visible candidate in $\mathcal{C}_t$.
Both \texttt{possible} and \texttt{confirmed} candidates are eligible for \text{\texttt{Approach}}, while \text{\texttt{Stop}} additionally requires a \texttt{confirmed} candidate within the policy's stopping distance.
A programmatic validator checks the returned decision against these constraints before execution.
When no candidate is available, the agent continues exploring.

\subsection{Fly Straight}
\label{sec:method_flight}

\paragraph{Local Waypoint Formation.}
To translate visual selections into executable motion while bounding travel per decision, we convert each validated movement action into a 3D waypoint $\bm{w}_t$:
\begin{itemize}[leftmargin=*,itemsep=1pt,topsep=1pt]
  \item For \textbf{$\text{\texttt{Explore}}_h$}, $\bm{w}_t$ lies one step away at the current altitude, along the horizontal projection of the calibrated ray through the selected region's center.
  
  \item For \textbf{$\text{\texttt{Explore}}_v$}, $\bm{w}_t$ lies directly above or below the UAV, enabling ascent or descent without horizontal translation.
  
  \item For \textbf{\text{\texttt{Approach}}}, $\bm{w}_t$ lies along the ray from the UAV toward the estimated surface point $\hat{\bm{p}}_{t,i}^{w}$ in Equation \ref{eq:candidate_grounding}, with travel constrained by the step limit and standoff distance.
\end{itemize}
An accepted \textbf{\texttt{Stop}} terminates the episode at the current position, requiring a currently visible candidate confirmed by the VLM to match the goal and valid depth localization within the policy’s stopping distance.

\paragraph{Straight-Line Flight.}
Each movement follows a bounded straight-line segment from $\bm{x}_t$ toward $\bm{w}_t$, with initial heading alignment for motion containing a horizontal component.
Before each short velocity command, the controller recomputes the direction to $\bm{w}_t$ from its measured position and checks the upcoming segment using local depth (Section \ref{sec:method_safety}), halting blocked motion independently of VLM inference.
Execution continues until the waypoint is reached or a local check interrupts the segment.
The measured endpoint, displacement, and termination status inform the next search decision.

\subsection{Check and Halt}
\label{sec:method_safety}

The path toward a selected waypoint may contain obstacles that require interrupting the motion. 
We therefore run local depth checks throughout execution, allowing the controller to halt without waiting for another VLM query.

\paragraph{Collision Check.}
Before each low-level movement command, the controller back-projects current depth measurements into 3D using camera calibration.
A 3D capsule represents the swept volume of the upcoming segment, with the planned segment as its axis and a radius equal to the UAV collision radius plus a safety margin.
If an observed obstacle point lies inside the capsule, the controller immediately blocks the command and halts the UAV at its current position.
This halt ends the current segment without issuing the task-level \texttt{Stop} action, allowing search to continue at the next decision step.

\paragraph{Execution Feedback and Closed-Loop State.}
Motion may end before the intended waypoint, so subsequent decisions must account for measured progress. 
We formalize action execution as:
\begin{equation}
    (\bm{x}^{\mathrm{end}}_t,f_t)
    = \operatorname{Exec}(a_t,\mathcal{O}_t;\mathcal{D}_t^{\mathrm{exec}}),
    \label{eq:execution_feedback}
\end{equation}
where $\mathcal{D}_t^{\mathrm{exec}}$ contains depth measurements acquired during motion.
The feedback $f_t$ records the measured endpoint $\bm{x}^{\mathrm{end}}_t$, net displacement $\|\bm{x}^{\mathrm{end}}_t-\bm{x}_t\|_2$, and termination status, such as completed, halted by obstacle, or command rejected.
These quantities allow subsequent planning to account for partial execution when motion ends before the intended waypoint. 
The program logs this feedback in history and passes it directly to the next decision, enabling the VLM to assess actual progress, reorient, or select another visual region after blocked motion.

\enlargethispage{\baselineskip}
\section{Experiments}
\label{sec:expt}

\begin{table*}[!t]
\setlength{\aboverulesep}{0pt}\setlength{\belowrulesep}{0pt}
\centering
\small
\renewcommand{\arraystretch}{1.4}
\caption{\textbf{Performance comparison on the UAV-ON benchmark across different target object sizes.}
\textsuperscript{\ensuremath{\diamondsuit}} OctMem-Agent was fine-tuned on UAV-ON from pretrained OpenVLA~\cite{kim2025openvla} weights.
\textsuperscript{\ensuremath{\clubsuit}} We reproduced these results. \textbf{Bold} and \underline{underlined} indicate the best and second-best results, respectively.}
\label{tab:uavon_main}
\vspace{-0.5em}
\makeatletter
\newcommand{\uavonheaderbackground}{%
  \noalign{%
    \dimen0=\dimexpr(\ht\@arstrutbox+\dp\@arstrutbox)*2+\cmidrulewidth\relax
    {\color{tbhead}\hrule height\dimen0}%
    \vskip-\dimen0}}
\makeatother
\resizebox{1.0\textwidth}{!}{%
\begin{tabular}{ll ccc ccc ccc ccc}
  \toprule
  \uavonheaderbackground
  & & \multicolumn{3}{c}{\textbf{Small}}
  & \multicolumn{3}{c}{\textbf{Medium}}
  & \multicolumn{3}{c}{\textbf{Large}}
  & \multicolumn{3}{c}{\textbf{Total}} \\
  \cmidrule(lr){3-5} \cmidrule(lr){6-8} \cmidrule(lr){9-11} \cmidrule(lr){12-14}
  \multirow{-2}{*}{\textbf{Method}} & \multirow{-2}{*}{\textbf{Model / Backbone}}
  & \textbf{SR}$\uparrow$ & \textbf{OSR}$\uparrow$ & \textbf{SPL}$\uparrow$
  & \textbf{SR}$\uparrow$ & \textbf{OSR}$\uparrow$ & \textbf{SPL}$\uparrow$
  & \textbf{SR}$\uparrow$ & \textbf{OSR}$\uparrow$ & \textbf{SPL}$\uparrow$
  & \textbf{SR}$\uparrow$ & \textbf{OSR}$\uparrow$ & \textbf{SPL}$\uparrow$ \\
  \midrule
  \textbf{Random} & --
  & 4.14 & 7.80 & 2.80 & 3.33 & 8.10 & 3.05 & 2.48 & 8.07 & 1.62 & 3.70 & 8.00 & 2.66 \\
  \textbf{CLIP-H} & CLIP ViT-B/16
  & 2.86 & 8.43 & 1.51 & 10.95 & 16.67 & 7.17 & 13.04 & 19.25 & 10.53 & 6.20 & 11.90 & 4.15 \\ 
  \textbf{AOA-F} & GPT-4o mini
  & 4.45 & 16.38 & 1.61 & 10.48 & 17.62 & 6.36 & 14.29 & 21.74 & 10.66 & 7.30 & 17.50 & 4.06 \\
  \textbf{AerialVLN} & --
  & 4.13 & 19.87 & 4.01 & 4.76 & 17.62 & 4.90 & 1.86 & 19.25 & 1.86 & 3.90 & 19.30 & 3.72 \\
  \textbf{NaVid} & Vicuna-7B
  & 10.02 & 29.55 & 4.41 & 11.43 & 26.44 & 5.95 & 17.39 & 30.73 & 15.03 & 11.50 & 29.10 & 6.44 \\
  \textbf{OpenFly} & OpenVLA-7B
  & 12.40 & 26.23 & 6.35 & 10.00 & 20.48 & 4.19 & 13.04 & 31.65 & 7.56 & 12.00 & 25.90 & 6.09 \\
  \textbf{AerialDojo\textsuperscript{\ensuremath{\clubsuit}}} & Qwen3-VL-8B
  & 6.47 & 8.04 & 6.47 & 4.32 & 6.79 & 4.32 & 8.26 & 10.74 & 8.26 & 6.29 & 8.21 & 6.29 \\
  \textbf{APEX} & --
  & -- & -- & -- & -- & -- & -- & -- & -- & -- & 13.33 & 20.00 & 10.14 \\
  \textbf{OctMem-Agent\textsuperscript{\ensuremath{\clubsuit}}} & LLaMA-2-7B\textsuperscript{\ensuremath{\diamondsuit}}
  & 18.89 & 24.89 & \underline{16.68} & 21.60 & 25.93 & 19.23 & 23.97 & 28.93 & 22.72 & 20.33 & 25.78 & 18.24 \\
  \textbf{AeroBelief} & Qwen3-VL-8B
  & 20.67 & 39.11 & 9.11 & 25.31 & 32.72 & 12.03 & 20.16 & 26.05 & 14.40 & 21.61 & 35.57 & 10.62 \\
  \textbf{SearchWorld} & --
  & 21.62 & 35.29 & 7.23 & 21.80 & 34.64 & 8.35 & 35.02 & 37.59 & 10.76 & 23.80 & 35.52 & 8.03 \\

  \midrule 
  \rowcolor{tbzebra} & Qwen3-VL-8B
  & \underline{32.22} & 45.78 & 16.59 & 48.15 & 53.70 & 35.08 & \underline{50.41} & 53.72 & \textbf{41.05} & \underline{38.74} & 48.84 & 24.71 \\
  \rowcolor{tbzebra} & Qwen3.5-4B
  & 28.89 & \underline{51.56} & 16.12 & \textbf{54.94} & \textbf{59.88} & \underline{38.31} & 48.76 & \underline{54.55} & 39.00 & 37.93 & \underline{53.89} & \underline{24.80} \\
  \rowcolor{tbzebra} \multirow{-3}{*}{\textbf{SimpleUAV (Ours)}} & Qwen3.6-35B-A3B
  & \textbf{34.00} & \textbf{52.44} & \textbf{20.93} & \underline{54.32} & \underline{59.26} & \textbf{43.81} & \textbf{52.89} & \textbf{61.16} & \underline{41.00} & \textbf{41.61} & \textbf{55.39} & \textbf{29.30} \\
\bottomrule
\end{tabular}}
\end{table*}

\subsection{Experimental Setup}
\label{sec:expt:setup}

\paragraph{Datasets.}
We evaluate SimpleUAV on two aerial object-search benchmarks and one aerial VLN benchmark.
\textbf{UAV-ON}~\cite{xiao2025uavon} covers diverse outdoor environments for open-world search, with targets described by category, size, and appearance.
We use \textbf{AerialDojo-200K}~\cite{feng2026aerialdojo}'s semantic goal setting across 42 diverse scenes, with a 3 m success radius versus UAV-ON's 20 m.
\textbf{TravelUAV}~\cite{wang2024openuav} tests instruction-guided continuous aerial navigation under L1 assistance.

\paragraph{Evaluation Metrics.}
On UAV-ON and AerialDojo-200K, we report Success Rate (\textbf{SR}), Oracle Success Rate (\textbf{OSR}), and Success-weighted Path Length (\textbf{SPL}).
On TravelUAV, we report SR for the Easy, Hard, and Full settings.
We additionally report Collision Rate (\textbf{CR}) in the UAV-ON component ablations. SR, OSR, SPL, and CR are percentages.
Appendix~\ref{app:metrics} provides metric definitions.

\paragraph{Baselines.}
On UAV-ON, we compare with zero-shot baselines~\cite{xiao2025uavon}, aerial VLN methods~\cite{liu2023aerialvln, zhang2024navid, gao2025openfly}, and specialized aerial search systems.
These include \textbf{OctMem-Agent}~\cite{zhou2026octmem} with adaptive octree memory, \textbf{APEX}~\cite{zhang2026apex} with semantic maps and a learned action policy, \textbf{AeroBelief}~\cite{xiao2026aerobelief} with dual-layer semantic-spatial belief maps, and \textbf{SearchWorld}~\cite{ji2026searchworld} with a learned world model and imagined rollouts.
On AerialDojo-200K, we include benchmark-reported open- and closed-source VLM results and reproduce \textbf{AerialDojo's reference framework}~\cite{feng2026aerialdojo} with Qwen3-VL-4B and Qwen3-VL-8B.
We also run the 8B framework on UAV-ON and OctMem-Agent on AerialDojo-200K for cross-benchmark comparisons, with Qwen3-VL-8B enabling same-backbone comparisons with SimpleUAV and AeroBelief on UAV-ON.
On TravelUAV, we compare with \textbf{NavFoM}~\cite{zhang2025navfom}, \textbf{TravelUAV}~\cite{wang2024openuav}, and \textbf{AirForesight}~\cite{liu2026airforesight}.

\paragraph{Implementation Details.}
We run frozen VLMs with vLLM on NVIDIA RTX PRO 6000 GPUs without task-specific training (settings and model IDs in Appendix \ref{app:implementation}).
On AerialDojo-200K, SimpleUAV and all our reproduced baselines share MoGe2-estimated depth (Appendix~\ref{app:aerialdojo_depth}).

\subsection{Main Results}
\label{sec:expt:main}

\paragraph{Performance Across Object-Search Benchmarks.}
As shown in Table~\ref{tab:uavon_main}, all three SimpleUAV variants exceed the strongest baseline SR across small, medium, and large targets and overall on UAV-ON.
The best variant in each size category improves SR by \textbf{12.38\%}, \textbf{29.63\%}, and \textbf{17.87\%}, respectively.
Even the compact 4B variant achieves 37.93\% overall SR, surpassing the strongest prior baseline by 14.13\%. 
On AerialDojo-200K, which covers more diverse environments and employs a stricter 3 m success radius than UAV-ON's 20 m, our default configuration improves SR, OSR, and SPL over the best previously reported results by \textbf{8.78\%}, \textbf{9.38\%}, and \textbf{4.18\%}, respectively (Table~\ref{tab:aerialdojo_main}).
These results confirm that strong aerial search is achievable with a small frozen VLM.

SPL reaches 29.30\% versus 18.24\% on UAV-ON and 7.01\% versus 2.83\% on AerialDojo-200K.
With only 40 decisions, SimpleUAV still achieves 32.88\% SR and 26.95\% SPL on UAV-ON (Section~\ref{sec:expt:ablation}), exceeding every baseline score in Table \ref{tab:uavon_main} for both metrics.
Together, these results demonstrate strong performance across target sizes and success criteria, combining higher search success and path efficiency with a clear advantage under a tight decision budget.

\begin{table}[!t]
\setlength{\aboverulesep}{0pt}\setlength{\belowrulesep}{0pt}
\centering
\small
\setlength{\tabcolsep}{5pt}
\renewcommand{\arraystretch}{1.4}
\caption{\textbf{Performance comparison on the AerialDojo-200K Base benchmark under the 3\,m success criterion.}
\textsuperscript{\ensuremath{\diamondsuit}} OctMem-Agent was fine-tuned on UAV-ON from pretrained OpenVLA~\cite{kim2025openvla} weights.
\textsuperscript{\ensuremath{\clubsuit}} We reproduced these results. \textbf{Bold} and \underline{underlined} indicate the best and second-best results, respectively.}
\label{tab:aerialdojo_main}
\vspace{-0.5em}
\resizebox{1.0\columnwidth}{!}{%
\begin{tabular}{ll ccc}
  \toprule
  \rowcolor{tbhead}
  \textbf{Method} & \textbf{Backbone} & \textbf{SR}$\uparrow$ & \textbf{OSR}$\uparrow$ & \textbf{SPL}$\uparrow$ \\
  \midrule
  \multirow{9}{*}{\textbf{AerialDojo}} 
   & Qwen3-VL-4B & 0.71 & 1.50 & 0.51 \\
   & InternVL3.5-8B & 0.16 & 3.41 & 0.11 \\
   & Ministral-3-8B & 0.17 & 1.28 & 0.09 \\
   & Phi-4-Multimodal & 0.13 & 0.60 & 0.09 \\
   & MiniCPM-V-4.6 & 0.19 & 1.35 & 0.15 \\
   & GPT-5.6-Sol & 2.81 & 3.06 & 2.20 \\
   & Grok-4.5 & 0.25 & 0.49 & 0.25 \\
   & Gemini-3.8-Flash & 2.40 & 3.68 & 2.37 \\
   & Claude-Opus-5 & 3.20 & 7.84 & 2.83 \\
  \midrule
  \rowcolor{tbzebra} & Qwen3-VL-8B & 0.75 & 0.99 & 0.73 \\
  \rowcolor{tbzebra} \multirow{-2}{*}{\textbf{AerialDojo\textsuperscript{\ensuremath{\clubsuit}}}} & Qwen3-VL-4B & 0.67 & 0.99 & 0.65 \\
  \midrule
  \textbf{OctMem-Agent\textsuperscript{\ensuremath{\clubsuit}}}
   & LLaMA-2-7B\textsuperscript{\ensuremath{\diamondsuit}} & 0.40 & 1.25 & 0.39 \\
  \midrule
  \rowcolor{tbzebra} & Qwen3-VL-8B & \underline{9.65} & \underline{13.88} & \underline{4.99} \\
  \rowcolor{tbzebra} \multirow{-2}{*}{\textbf{SimpleUAV (Ours)}} & Qwen3.6-35B-A3B & \textbf{11.98} & \textbf{17.22} & \textbf{7.01} \\
  \bottomrule
\end{tabular}}
\end{table}

\paragraph{Cross-Benchmark Agent Comparisons.}
Across UAV-ON and AerialDojo-200K, AerialDojo's Qwen3-VL-8B reference framework achieves 6.29\% and 0.75\% SR, respectively.
With the identical backbone, SimpleUAV reaches \textbf{38.74\%} and \textbf{9.65\%}, delivering SR gains of \textbf{32.45\%} and \textbf{8.90\%}.
OctMem-Agent, fine-tuned on UAV-ON, drops sharply from 20.33\% SR on UAV-ON to 0.40\% on AerialDojo-200K.
These large same-backbone gains establish that SimpleUAV's advantage comes from its closed-loop agent design, which couples VLM reasoning with local waypoint execution and execution feedback.
The advantage persists across different environments and the 20\,m versus 3\,m success criteria, demonstrating strong cross-benchmark generalization of the same core agent without task-specific training or structural redesign.

\subsection{Transfer to Aerial VLN}
\label{sec:expt:transfer}

Following standard benchmark evaluation protocols, we evaluate transfer from object-goal search to aerial VLN on TravelUAV under L1 assistance, without additional training or fine-tuning.
Table~\ref{tab:traveluav_main} shows that SimpleUAV consistently exceeds AirForesight~\cite{liu2026airforesight} in Full SR by \textbf{21.10\%}, \textbf{19.66\%}, and \textbf{21.18\%} on the Seen, Unseen Maps (UM), and Unseen Objects (UO) splits, respectively.
It also achieves higher Easy and Hard SR across all splits, supporting transfer to instruction-guided aerial navigation.

\begin{table}[!t]
\setlength{\aboverulesep}{0pt}\setlength{\belowrulesep}{0pt}
\centering
\small
\setlength{\tabcolsep}{4pt}
\renewcommand{\arraystretch}{1.4}
\caption{\textbf{Success rate (\%) on TravelUAV under L1 assistance.}}
\label{tab:traveluav_main}
\vspace{-0.5em}
\resizebox{0.99\columnwidth}{!}{%
\begin{tabular}{ll ccc}
  \toprule
  \rowcolor{tbhead}
  \textbf{Split} & \textbf{Method} & \textbf{Easy SR}$\uparrow$ & \textbf{Hard SR}$\uparrow$ & \textbf{Full SR}$\uparrow$ \\ 
  \midrule
  & \textbf{NavFoM} & 32.91 & 23.58 & 29.17 \\
  & \textbf{TravelUAV} & 32.40 & 28.03 & 30.47 \\
  & \textbf{AirForesight} & 38.83 & 32.13 & 35.83 \\
  \multirow{-4}{*}{\textbf{Seen}} & \textbf{SimpleUAV (Ours)} & \textbf{60.52} & \textbf{52.49} & \textbf{56.93} \\
  \midrule
  \rowcolor{tbzebra} & \textbf{NavFoM} & 6.77 & 5.36 & 6.30 \\
  \rowcolor{tbzebra} & \textbf{TravelUAV} & 11.23 & 8.06 & 9.92 \\
  \rowcolor{tbzebra} & \textbf{AirForesight} & 15.86 & 10.58 & 13.67 \\
  \rowcolor{tbzebra} \multirow{-4}{*}{\textbf{Unseen Maps}} & \textbf{SimpleUAV (Ours)} & \textbf{38.50} & \textbf{26.01} & \textbf{33.33} \\
  \midrule
  & \textbf{NavFoM} & 32.54 & 28.03 & 29.83 \\
  & \textbf{TravelUAV} & 42.11 & 39.09 & 40.10 \\
  & \textbf{AirForesight} & 48.80 & 42.14 & 44.36 \\
  \multirow{-4}{*}{\textbf{Unseen Objects}} & \textbf{SimpleUAV (Ours)} & \textbf{75.73} & \textbf{60.48} & \textbf{65.54} \\
  \bottomrule
\end{tabular}}
\end{table}

\subsection{Ablation Study}
\label{sec:expt:ablation}

\paragraph{Effect of VLM Backbones.}
Figure~\ref{fig:model_comparison} compares backbone performance under 40- and 150-decision budgets.
The gap between compact 4B models and larger backbones narrows at 150 decisions, where both 4B variants exceed the strongest prior system by over \textbf{11\%} in SR.
Additional decisions may give compact models more opportunity to recover from suboptimal actions.
Parameter count alone does not determine performance, as Qwen3.5-9B outperforms the three-times-larger Qwen3.8-27B under 40 decisions.
At 150 decisions, the evaluated 8B and 9B dense models attain the same SR, while Qwen3.6-35B-A3B achieves the highest SR of \textbf{41.61\%} with 3B active parameters per token, motivating its use as our \emph{default} backbone.

\begin{figure}[!t]
  \centering
  \includegraphics[width=\columnwidth]{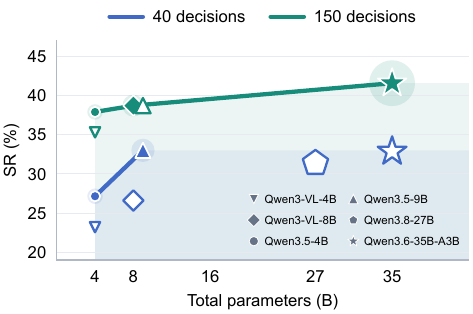}
  \vspace{-1.5em}
  \caption{\textbf{Final SR versus total backbone parameters on UAV-ON with 40/150 decisions.} Marker shape identifies the backbone, and marker area is proportional to total parameters. Filled markers and connecting lines trace each budget's Pareto frontier.}
  \label{fig:model_comparison}
  \vspace{-0.5em}
\end{figure}

\begin{figure}[!t]
  \centering
  \includegraphics[width=\columnwidth]{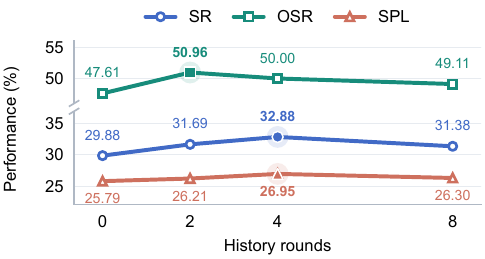}
  \vspace{-1.5em}
  \caption{\textbf{Effect of interaction history length on UAV-ON using Qwen3.6-35B-A3B under a 40-decision budget.} The Y-axis break omits 37--46\%.}
  \label{fig:history_ablation_40}
  \vspace{-0.5em}
\end{figure}

\paragraph{Effect of Interaction History.}
Figure~\ref{fig:history_ablation_40} depicts that under a 40-decision budget, the 0-round variant achieves 29.88\% SR, only 3\% below the best setting.
Two rounds maximize OSR at 50.96\%, compared with 47.61\% at zero rounds, while four rounds maximize SR and SPL at 32.88\% and 26.95\%.
Extending history to eight rounds reduces all three metrics relative to four rounds, yielding 31.38\% SR, 49.11\% OSR, and 26.30\% SPL.
These results favor compact recent context for action selection.
Longer histories may introduce outdated observations while increasing input length, so we retain \emph{four rounds} by default.

\begin{figure}[!t]
  \centering
  \includegraphics[width=\columnwidth]{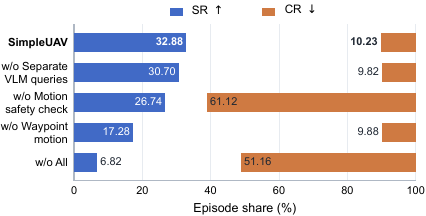}
  \vspace{-1.5em}
  \caption{\textbf{Component ablations on UAV-ON using Qwen3.6-35B-A3B under a 40-decision budget.} Blank gaps denote episodes with neither success nor collision.}
  \label{fig:motion_ablation}
\end{figure}

\paragraph{Effect of Separate VLM Queries.}
Merging target assessment and action selection into one call reduces SR from 32.88\% to 30.70\% (\textit{w/o Separate VLM Queries}, Figure~\ref{fig:motion_ablation}).
This \textbf{2.18\%} gap demonstrates the benefit of dedicated current-view verification: target judgments rely on fresh observations, while history-conditioned planning maintains search continuity.

\paragraph{Effect of Motion Safety Checking.}
As displayed in Figure~\ref{fig:motion_ablation}, removing reactive depth checks during straight-line execution reduces SR from 32.88\% to 26.74\% and increases CR from 10.23\% to 61.12\% under a 40-decision budget.
These results support checking geometry throughout execution.
The capsule test evaluates the upcoming swept volume before each command, allowing the controller to reject detected unsafe movements and return obstacle-induced termination status to the next VLM decision.

\paragraph{Effect of Waypoint Motion.}
Under the same 40-decision budget, replacing image-region-conditioned 3D waypoints with eight fixed action primitives while retaining bounding boxes and interaction history reduces SR from 32.88\% to 17.28\% (Figure~\ref{fig:motion_ablation}).
Fixed translation and rotation increments restrict how movement direction and distance can adapt to selected regions and local geometry.
This result supports grounding visual selections in 3D waypoints for exploration and target approach.

\paragraph{A Minimal Fixed-Action Baseline.}
The \textit{w/o All} baseline receives only current RGB observations and the target description, directly selecting among eight fixed action primitives without depth, interaction history, or collision checks (Figure~\ref{fig:motion_ablation}).
It achieves 6.82\% SR under the same 40-decision budget.
The \textbf{26.06\%} SR gap quantifies the combined benefit of SimpleUAV's geometric grounding, motion checks, and search context over this minimal baseline.
Together with the preceding ablations, this comparison supports integrating these components into the search loop.

\begin{figure}[!t]
  \centering
  \includegraphics[width=0.99\linewidth]{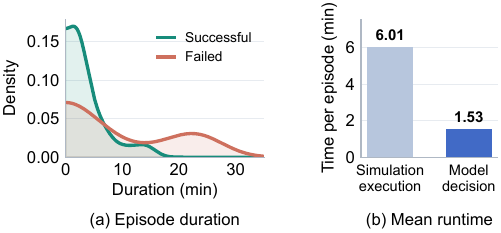}
  \vspace{-0.5em}
  \caption{\textbf{Runtime profile.}
  (a) Duration densities for successful and failed episodes. (b) Mean runtime components per episode over all outcomes. Simulation and execution time is total elapsed time minus recorded model inference and grounding time.}
  \label{fig:runtime_analysis}
  \vspace{-0.5em}
\end{figure}

\paragraph{Efficiency Analysis.}
Figure~\ref{fig:runtime_analysis}(a) shows the episode-duration distributions for successful and failed searches.
Model inference and grounding average 1.53\,min per episode, accounting for 20.3\% of total elapsed time. Simulation and other execution overhead average 6.01\,min per episode and account for the remaining 79.7\% (Figure~\ref{fig:runtime_analysis}(b)).
The runtime is therefore dominated by simulation and execution, with model decision-making taking a smaller share.

\paragraph{Why Does SimpleUAV Work?}
The experimental results support coordinating semantic decisions, bounded search context, and local geometric execution.
With Qwen3-VL-8B, SimpleUAV exceeds AeroBelief on UAV-ON (38.74\% versus 21.61\% SR) and AerialDojo's reference framework on both object-search benchmarks. These comparisons show that navigation design materially affects performance with a fixed backbone.
In SimpleUAV, the VLM assesses current candidates and selects actions using recent context, while local geometry handles candidate localization, waypoint formation, bounded motion, and capsule collision checks. Measured execution feedback connects successive decisions.
The waypoint and safety ablations support the geometric execution components, while the history study favors compact context.
Strong results with multiple frozen backbones, including 4B models, further show that the design is effective across the evaluated model choices.

\subsection{Case Study}
\label{sec:expt:case}

\begin{figure}[!t]
  \centering
  \includegraphics[width=\linewidth]{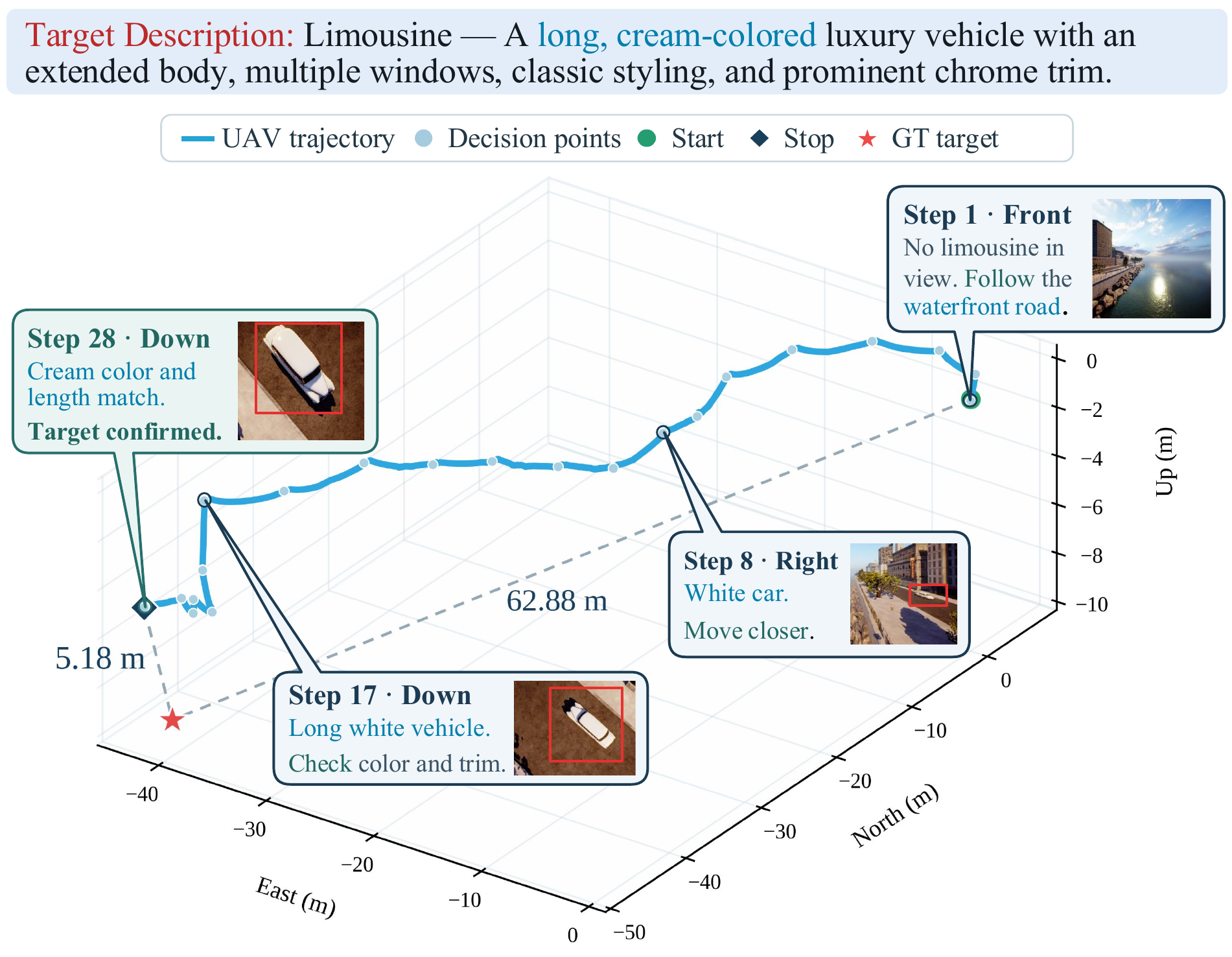}
  \vspace{-1.5em}
  \caption{\textbf{SimpleUAV searching for a limousine.}
  The trajectory and selected views illustrate exploration, candidate inspection, and confirmation.
  The accompanying textual reports document the observed attributes, inspection criteria, and action choices throughout the episode.
  }
  \label{fig:case_study}
\end{figure}

Figure~\ref{fig:case_study} illustrates a limousine-search episode starting 62.88 m from the target.
With no target initially visible, the agent follows the waterfront road (Step 1).
A white vehicle appears in the right view at Step 8, prompting an approach for closer inspection.
At Step 17, the downward view reveals its elongated body, motivating further examination of color and trim.
At Step 28, the current view supports a confirmed match, and the UAV stops 5.18 m from the ground-truth target.
This sequence illustrates how successive approaches provide new views for target assessment before stopping.
Search context guides subsequent actions, while candidate identity is reassessed from current observations at each decision.

\section{Conclusions}
\label{sec:conclusions}


We introduce \textbf{SimpleUAV}, demonstrating that a minimal agent can outperform specialized aerial navigation systems without task-specific training. 
SimpleUAV separates current-view target assessment from history-conditioned action selection and connects bounded waypoint motion, local depth checks, and measured execution feedback in a closed loop. 
It achieves the highest success rates among compared methods across two object-search benchmarks and aerial VLN, with limited benchmark-specific adaptation.
These results support a practical design principle: combine semantic reasoning and compact search context with local geometric execution, without requiring persistent spatial maps or learned world models. 
Extending this design to broader embodied search and incorporating lightweight memory for global coverage are promising directions. 
SimpleUAV establishes a strong empirical baseline for determining when additional memory, predictive modeling, or policy training delivers measurable benefits.

\section*{LLM Usage Statement}
OpenAI ChatGPT and Codex assisted with literature investigation, manuscript drafting, and experimental code. The authors are responsible for all claims and results.

\bibliographystyle{ieeenat_fullname}
\bibliography{references}

@String(CVPR = {CVPR})

@String(ICCV = {ICCV})

@String(ECCV = {ECCV})

@String(NIPS = {NeurIPS})

@String(ACMMM = {ACM MM})

@String(ICLR = {ICLR})

@String(PR = {PR})

@String(AAAI = {AAAI})

@String(ACL = {ACL})

@String(EMNLP = {EMNLP})

@String(ICRA = {ICRA})

@String(IROS = {IROS})

@String(CORL = {CoRL})

@String(CIRA = {CIRA})

@String(RAL = {RA-L})

@String(TVT = {TVT})

@String(TIM = {TIM})

@String(PMLR = {PMLR})

@InProceedings{xiao2025uavon,
  author    = {Xiao, Jianqiang and Sun, Yuexuan and Shao, Yixin and Gan, Boxi and Liu, Rongqiang and Wu, Yanjin and Guan, Weili and Deng, Xiang},
  booktitle = ACMMM,
  title     = {{UAV-ON}: A benchmark for open-world object goal navigation with aerial agents},
  year      = {2025},
  pages     = {13023--13029},
  url       = {https://dl.acm.org/doi/10.1145/3746027.3758251},
}

@InProceedings{ji2026cityavos,
  author    = {Ji, Yatai and Zhu, Zhengqiu and Zhao, Yong and Liu, Beidan and Gao, Chen and Zhao, Yihao and Qiu, Sihang and Hu, Yue and Yin, Quanjun},
  booktitle = AAAI,
  title     = {Towards autonomous {UAV} visual object search in city space: Benchmark and agentic methodology},
  year      = {2026},
  number    = {22},
  pages     = {18342--18350},
  volume    = {40},
  url       = {https://ojs.aaai.org/index.php/AAAI/article/view/38898},
}

@inproceedings{zhang2026apex,
  author = {Daoxuan Zhang and Ping Chen and Xiaobo Xia and Xiu Su and Ruichen Zhen and Jianqiang Xiao and Shuo Yang},
  title = {{APEX}: A Decoupled Memory-based Explorer for Asynchronous Aerial Object Goal Navigation},
  booktitle = CVPR,
  pages = {15232--15242},
  year = {2026},
  url = {https://openaccess.thecvf.com/content/CVPR2026/html/Zhang_APEX_A_Decoupled_Memory-based_Explorer_for_Asynchronous_Aerial_Object_Goal_CVPR_2026_paper.html}
}

@InProceedings{zhou2026octmem,
  author    = {Zhou, Jiacong and Miao, Jiaxu and Lin, Yourun and Wang, Xianyun and Xiao, Jun and Yu, Jun},
  booktitle = CVPR,
  title     = {Memory-Augmented Scene Understanding and Exploration for Open-World Aerial Object-Goal Navigation},
  year      = {2026},
  pages     = {21616--21626},
  url       = {https://openaccess.thecvf.com/content/CVPR2026/html/Zhou_Memory-Augmented_Scene_Understanding_and_Exploration_for_Open-World_Aerial_Object-Goal_Navigation_CVPR_2026_paper.html},
}

@InProceedings{kim2025openvla,
  title = 	 {OpenVLA: An Open-Source Vision-Language-Action Model},
  author =       {Kim, Moo Jin and Pertsch, Karl and Karamcheti, Siddharth and Xiao, Ted and Balakrishna, Ashwin and Nair, Suraj and Rafailov, Rafael and Foster, Ethan P and Sanketi, Pannag R and Vuong, Quan and Kollar, Thomas and Burchfiel, Benjamin and Tedrake, Russ and Sadigh, Dorsa and Levine, Sergey and Liang, Percy and Finn, Chelsea},
  booktitle = 	 CoRL,
  pages = 	 {2679--2713},
  year = 	 {2025},
  volume = 	 {270},
  series = 	 {Proceedings of Machine Learning Research},
  url = 	 {https://proceedings.mlr.press/v270/kim25c.html},
}

@InProceedings{kim2026raven,
  author       = {Kim, Seungchan and Alama, Omar and Kurdydyk, Dmytro and Keller, John and Keetha, Nikhil and Wang, Wenshan and Bisk, Yonatan and Scherer, Sebastian},
  booktitle    = ICRA,
  title        = {{RAVEN}: Resilient aerial navigation via open-set semantic memory and behavior adaptation},
  year         = {2026},
  organization = {IEEE},
  pages        = {3808--3815},
  url          = {https://ieeexplore.ieee.org/document/11696892},
}

@ARTICLE{chen2026airhunt,
  author={Chen, Xuecheng and Liu, Zongzhuo and Ma, Jianfa and Du, Bang and Zhang, Tiantian and Wang, Xueqian and Zhou, Boyu},
  journal={IEEE Robotics and Automation Letters},
  title={AirHunt: Bridging VLM Semantics and Continuous Planning for Efficient Aerial Object Navigation},
  year={2026},
  volume={11},
  number={11},
  pages={12495-12502},
  url = {https://ieeexplore.ieee.org/document/11676199},
}

@misc{ji2026searchworld,
      title={SearchWorld: Spatial Value-Grounded Imagination for UAV Object Search via World Models},
      author={Yatai Ji and Zhengqiu Zhu and Yong Zhao and Yue Hu and Fanglong Yao and Chen Gao and Pengfei Zhu and Quanjun Yin},
      year={2026},
      eprint={2610.09335},
      archivePrefix={arXiv},
      primaryClass={cs.AI},
      url={https://arxiv.org/abs/2610.09335},
}

@Article{xiao2026aerobelief,
  author  = {Xiao, Jianqiang and Deng, Xiang and Sun, Yuexuan and Wu, Yanjin and Yan, Wenbiao and Nie, Liqiang},
  journal = {arXiv preprint arXiv:2609.08164},
  title   = {Dual-Layer Semantic-Spatial Belief Mapping for Aerial Object Goal Navigation},
  year    = {2026},
  url     = {https://arxiv.org/abs/2609.08164},
}

@article{yang2026skyagent,
  author = {Kaibo Yang and Xingyu Wu and Enze Zhu and Yajie Chai and Xuyang Li and Jiahui Cui and Xiaoxuan Liu and Lei Wang and Zhan Chen},
  title = {{SkyAgent}: Revisable Target Memory for Continuous Aerial Search},
  journal = {arXiv preprint arXiv:2610.23293},
  year = {2026},
  url = {https://arxiv.org/abs/2610.23293}
}

@article{zhang2024navid,
        title={NaVid: Video-based VLM Plans the Next Step for Vision-and-Language Navigation},
        author={Zhang, Jiazhao and Wang, Kunyu and Xu, Rongtao and Zhou, Gengze and Hong, Yicong and Fang, Xiaomeng and Wu, Qi and Zhang, Zhizheng and Wang, He},
        journal={Robotics: Science and Systems},
        year={2024},
        url= {https://www.roboticsproceedings.org/rss20/p079.html},
}

@InProceedings{zhou2024navgpt2,
  author       = {Zhou, Gengze and Hong, Yicong and Wang, Zun and Wang, Xin Eric and Wu, Qi},
  booktitle    = ECCV,
  title        = {{NavGPT-2}: Unleashing navigational reasoning capability for large vision-language models},
  year         = {2024},
  organization = {Springer},
  pages        = {260--278},
  url          = {https://link.springer.com/chapter/10.1007/978-3-031-72667-5_15},
}

@Article{gao2025stmr,
  author  = {Gao, Yunpeng and Wang, Zhigang and Han, Pengfei and Jing, Linglin and Wang, Dong and Zhao, Bin},
  journal = {arXiv preprint arXiv:2410.08500},
  title   = {Exploring spatial representation to enhance {LLM} reasoning in aerial vision-language navigation},
  year    = {2024},
  url     = {https://arxiv.org/abs/2410.08500},
}

@InProceedings{zhang2025citynavagent,
  author    = {Zhang, Weichen and Gao, Chen and Yu, Shiquan and Peng, Ruiying and Zhao, Baining and Zhang, Qian and Cui, Jinqiang and Chen, Xinlei and Li, Yong},
  booktitle = ACL,
  title     = {{CityNavAgent}: Aerial vision-and-language navigation with hierarchical semantic planning and global memory},
  year      = {2025},
  pages     = {31292--31309},
  url       = {https://aclanthology.org/2025.acl-long.1511/},
}

@InProceedings{wang2024openuav,
  author    = {Wang, Xiangyu and Yang, Donglin and Wang, Ziqin and Kwan, Hohin and Chen, Jinyu and Li, Hongsheng and Liao, Yue and Liu, Si and others},
  booktitle = ICLR,
  title     = {Towards realistic {UAV} vision-language navigation: Platform, benchmark, and methodology},
  year      = {2025},
  pages     = {7292--7310},
  volume    = {2025},
  url       = {https://proceedings.iclr.cc/paper_files/paper/2025/hash/15ce8e7afe5ee95bad56e3b9be28d3d1-Abstract-Conference.html},
}

@inproceedings{gao2025openfly,
 author = {Gao, Yunpeng and Li, Chenhui and You, Zhongrui and Liu, Junli and Zhen, Li and CHEN, Pengan and Chen, Qizhi and Tang, Zhonghan and Wang, Liansheng and Yangpenghui and Tang, Yiwen and Tang, Yuhang and Liang, Shuai and Zhu, Songyi and Xiong, Ziqin and Su, Yifei and Ye, Xinyi and Li, Jianan and Ding, Yan and Wang, Dong and Wang, Zhigang and Zhao, Bin and Li, Xuelong},
 booktitle = ICLR,
 title = {OpenFly: A COMPREHENSIVE PLATFORM FOR AERIAL VISION-LANGUAGE NAVIGATION},
 year = {2026},
 url = {https://proceedings.iclr.cc/paper_files/paper/2026/file/a6fcfd15cd01e4a550808c3e01f5583d-Paper-Conference.pdf}
}

@Article{jiang2025longfly,
  author  = {Jiang, Wen and Wang, Li and Huang, Kangyao and Fan, Wei and Liu, Jinyuan and Liu, Shaoyu and Duan, Hongwei and Xu, Bin and Ji, Xiangyang},
  journal = {arXiv preprint arXiv:2512.22010},
  title   = {{LongFly}: Long-horizon {UAV} vision-and-language navigation with spatiotemporal context integration},
  year    = {2025},
  url     = {https://arxiv.org/abs/2512.22010},
}

@Misc{jiang2026spatialfly,
  author        = {Wen Jiang and Kangyao Huang and Li Wang and Wang Xu and Wei Fan and Jinyuan Liu and Shaoyu Liu and Hanfang Liang and Hongwei Duan and Bin Xu and Xiangyang Ji and Huaping Liu},
  title         = {{SpatialFly}: Implicit {3D} Prior-Guided Visual Reparameterization for Continuous {UAV} Vision-and-Language Navigation},
  year          = {2026},
  archiveprefix = {arXiv},
  eprint        = {2603.21046},
  howpublished  = {arXiv preprint arXiv:2603.21046},
  primaryclass  = {cs.CV},
  url           = {https://arxiv.org/abs/2603.21046},
}

@Article{wang2026forefly,
  author  = {Wang, Kunhui and Zhang, Xintong and Gao, Junyu and Xu, Changsheng},
  journal = {arXiv preprint arXiv:2609.33581},
  title   = {{ForeFly}: A Dual-Horizon World Action Model for Aerial Vision-Language Navigation},
  year    = {2026},
  url     = {https://arxiv.org/abs/2609.33581},
}

@InProceedings{liu2023aerialvln,
  author       = {Liu, Shubo and Zhang, Hongsheng and Qi, Yuankai and Wang, Peng and Zhang, Yanning and Wu, Qi},
  booktitle    = ICCV,
  title        = {{AerialVLN}: Vision-and-language navigation for {UAVs}},
  year         = {2023},
  organization = {IEEE},
  pages        = {15384--15394},
  url          = {https://openaccess.thecvf.com/content/ICCV2023/html/Liu_AerialVLN_Vision-and-Language_Navigation_for_UAVs_ICCV_2023_paper.html},
}

@article{feng2026aerialdojo,
  author  = {Feng, Tongtong and Wang, Xin and Hou, Haoran and Wang, Ren and Wang, Weiran and Zhu, Shaokai and Jia, Ziqi and Wang, Hao and Zhan, Yu-Wei and Wu, Zongyuan and Cui, Jinghao and Zhu, Wenwu},
  title   = {{AerialDojo-200K}: A Large-Scale Benchmark Suite for Open-World Aerial Object-Goal Search},
  journal = {arXiv preprint arXiv:2609.36066},
  year    = {2026},
  url     = {https://arxiv.org/abs/2609.36066},
}

@article{zhang2025navfom,
  author  = {Zhang, Jiazhao and Li, Anqi and Qi, Yunpeng and Li, Minghan and Liu, Jiahang and Wang, Shaoan and Liu, Haoran and Zhou, Gengze and Wu, Yuze and Li, Xingxing and Fan, Yuxin and Li, Wenjun and Chen, Zhibo and Gao, Fei and Wu, Qi and Zhang, Zhizheng and Wang, He},
  title   = {Embodied Navigation Foundation Model},
  journal = {arXiv preprint arXiv:2509.12129},
  year    = {2025},
  url     = {https://arxiv.org/abs/2509.12129},
}

@inproceedings{liu2026airforesight,
  author    = {Liu, Yutong and Li, Xiaojie and Xu, Mingzhu and Wu, Jianlong},
  title     = {{AirForesight}: Current-to-Future Spatial Map Imagination with Cross-Space Planning Consistency for {UAV-VLN}},
  booktitle = ACMMM,
  year      = {2026},
  doi       = {10.1145/3767308.3836065},
  url       = {https://doi.org/10.1145/3767308.3836065},
}

@article{bai2025qwen3,
  title   = {{Qwen3-VL} Technical Report},
  author  = {Bai, Shuai and Cai, Yuxuan and Chen, Ruizhe and Chen, Keqin and Chen, Xionghui and Cheng, Zesen and Deng, Lianghao and Ding, Wei and Gao, Chang and Ge, Chunjiang and others},
  journal = {arXiv preprint arXiv:2511.21631},
  year    = {2025},
  url     = {https://arxiv.org/abs/2511.21631},
}

@misc{qwen36_35b_a3b,
  title  = {{Qwen3.6-35B-A3B}: Agentic Coding Power, Now Open to All},
  author = {{Qwen Team}},
  month  = {April},
  year   = {2026},
  url    = {https://qwen.ai/blog?id=qwen3.6-35b-a3b},
}

@inproceedings{anderson2018r2r,
  author    = {Anderson, Peter and Wu, Qi and Teney, Damien and Bruce, Jake and Johnson, Mark and S{\"u}nderhauf, Niko and Reid, Ian and Gould, Stephen and van den Hengel, Anton},
  title     = {Vision-and-Language Navigation: Interpreting Visually-Grounded Navigation Instructions in Real Environments},
  booktitle = CVPR,
  year      = {2018},
  pages     = {3674--3683},
  url       = {https://openaccess.thecvf.com/content_cvpr_2018/html/Anderson_Vision-and-Language_Navigation_Interpreting_CVPR_2018_paper.html},
}

@inproceedings{fried2018speaker,
  author    = {Fried, Daniel and Hu, Ronghang and Cirik, Volkan and Rohrbach, Anna and Andreas, Jacob and Morency, Louis-Philippe and Berg-Kirkpatrick, Taylor and Saenko, Kate and Klein, Dan and Darrell, Trevor},
  title     = {Speaker-Follower Models for Vision-and-Language Navigation},
  booktitle = NIPS,
  volume    = {31},
  year      = {2018},
  url       = {https://proceedings.neurips.cc/paper_files/paper/2018/hash/6a81681a7af700c6385d36577ebec359-Abstract.html},
}

@inproceedings{ku2020rxr,
  author    = {Ku, Alexander and Anderson, Peter and Patel, Roma and Ie, Eugene and Baldridge, Jason},
  title     = {Room-Across-Room: Multilingual Vision-and-Language Navigation with Dense Spatiotemporal Grounding},
  booktitle = EMNLP,
  year      = {2020},
  pages     = {4392--4412},
  doi       = {10.18653/v1/2020.emnlp-main.356},
  url       = {https://aclanthology.org/2020.emnlp-main.356/},
}

@inproceedings{krantz2020vlnce,
  author    = {Krantz, Jacob and Wijmans, Erik and Majumdar, Arjun and Batra, Dhruv and Lee, Stefan},
  title     = {Beyond the Nav-Graph: Vision-and-Language Navigation in Continuous Environments},
  booktitle = ECCV,
  year      = {2020},
  doi       = {10.1007/978-3-030-58604-1_7},
  url       = {https://www.ecva.net/papers/eccv_2020/papers_ECCV/papers/123730103.pdf},
}

@inproceedings{chen2021hamt,
  author    = {Chen, Shizhe and Guhur, Pierre-Louis and Schmid, Cordelia and Laptev, Ivan},
  title     = {History Aware Multimodal Transformer for Vision-and-Language Navigation},
  booktitle = NIPS,
  volume    = {34},
  year      = {2021},
  url       = {https://proceedings.neurips.cc/paper/2021/hash/2e5c2cb8d13e8fba78d95211440ba326-Abstract.html},
}

@inproceedings{zhou2024navgpt,
  author    = {Zhou, Gengze and Hong, Yicong and Wu, Qi},
  title     = {{NavGPT}: Explicit Reasoning in Vision-and-Language Navigation with Large Language Models},
  booktitle = AAAI,
  volume    = {38},
  number    = {7},
  pages     = {7641--7649},
  year      = {2024},
  doi       = {10.1609/aaai.v38i7.28597},
  url       = {https://ojs.aaai.org/index.php/AAAI/article/view/28597},
}

@inproceedings{chen2024mapgpt,
  author    = {Chen, Jiaqi and Lin, Bingqian and Xu, Ran and Chai, Zhenhua and Liang, Xiaodan and Wong, Kwan-Yee},
  title     = {{MapGPT}: Map-Guided Prompting with Adaptive Path Planning for Vision-and-Language Navigation},
  booktitle = ACL,
  year      = {2024},
  pages     = {9796--9810},
  doi       = {10.18653/v1/2024.acl-long.529},
  url       = {https://aclanthology.org/2024.acl-long.529/},
}

@inproceedings{long2024discussnav,
  author    = {Long, Yuxing and Li, Xiaoqi and Cai, Wenzhe and Dong, Hao},
  title     = {Discuss Before Moving: Visual Language Navigation via Multi-expert Discussions},
  booktitle = ICRA,
  year      = {2024},
  pages     = {17380--17387},
  doi       = {10.1109/ICRA57147.2024.10611565},
  url       = {https://doi.org/10.1109/ICRA57147.2024.10611565},
}

@inproceedings{ichter2023saycan,
  author    = {ichter, brian and Brohan, Anthony and Chebotar, Yevgen and Finn, Chelsea and Hausman, Karol and Herzog, Alexander and Ho, Daniel and Ibarz, Julian and Irpan, Alex and Jang, Eric and Julian, Ryan and Kalashnikov, Dmitry and Levine, Sergey and Lu, Yao and Parada, Carolina and Rao, Kanishka and Sermanet, Pierre and Toshev, Alexander T and Vanhoucke, Vincent and Xia, Fei and Xiao, Ted and Xu, Peng and Yan, Mengyuan and Brown, Noah and Ahn, Michael and Cortes, Omar and Sievers, Nicolas and Tan, Clayton and Xu, Sichun and Reyes, Diego and Rettinghouse, Jarek and Quiambao, Jornell and Pastor, Peter and Luu, Linda and Lee, Kuang-Huei and Kuang, Yuheng and Jesmonth, Sally and Joshi, Nikhil J. and Jeffrey, Kyle and Ruano, Rosario Jauregui and Hsu, Jasmine and Gopalakrishnan, Keerthana and David, Byron and Zeng, Andy and Fu, Chuyuan Kelly},
  title     = {Do As I Can, Not As I Say: Grounding Language in Robotic Affordances},
  booktitle = CORL,
  series    = PMLR,
  volume    = {205},
  pages     = {287--318},
  year      = {2023},
  url       = {https://proceedings.mlr.press/v205/ichter23a.html},
}

@inproceedings{huang2023innermonologue,
  author    = {Huang, Wenlong and Xia, Fei and Xiao, Ted and Chan, Harris and Liang, Jacky and Florence, Pete and Zeng, Andy and Tompson, Jonathan and Mordatch, Igor and Chebotar, Yevgen and Sermanet, Pierre and Jackson, Tomas and Brown, Noah and Luu, Linda and Levine, Sergey and Hausman, Karol and ichter, brian},
  title     = {Inner Monologue: Embodied Reasoning through Planning with Language Models},
  booktitle = CORL,
  series    = PMLR,
  volume    = {205},
  pages     = {1769--1782},
  year      = {2023},
  url       = {https://proceedings.mlr.press/v205/huang23c.html},
}

@inproceedings{shah2023lmnav,
  author    = {Shah, Dhruv and Osi{\'n}ski, B{\l}a{\.z}ej and ichter, brian and Levine, Sergey},
  title     = {{LM-Nav}: Robotic Navigation with Large Pre-Trained Models of Language, Vision, and Action},
  booktitle = CORL,
  series    = PMLR,
  volume    = {205},
  pages     = {492--504},
  year      = {2023},
  url       = {https://proceedings.mlr.press/v205/shah23b.html},
}

@inproceedings{qi2026pscavdn,
  author    = {Qi, Yu and Li, Hongyu and Huang, Shaofei and Hui, Tianrui and Wang, Yaxiong and Cheng, Lechao and Zhong, Zhun and Liu, Si and Wang, Meng},
  title     = {Parse, Search, and Confirmation: Training-Free Aerial Vision-and-Dialog Navigation with Chain-of-Thought Reasoning and Structured Spatial Memory},
  booktitle = CVPR,
  year      = {2026},
  pages     = {23859--23868},
  url       = {https://openaccess.thecvf.com/content/CVPR2026/html/Qi_Parse_Search_and_Confirmation_Training-Free_Aerial_Vision-and-Dialog_Navigation_with_Chain-of-Thought_CVPR_2026_paper.html},
}

@inproceedings{chaplot2020semexp,
  author    = {Chaplot, Devendra Singh and Gandhi, Dhiraj Prakashchand and Gupta, Abhinav and Salakhutdinov, Ruslan},
  title     = {Object Goal Navigation using Goal-Oriented Semantic Exploration},
  booktitle = NIPS,
  volume    = {33},
  pages     = {4247--4258},
  year      = {2020},
  url       = {https://proceedings.neurips.cc/paper/2020/hash/2c75cf2681788adaca63aa95ae028b22-Abstract.html},
}

@inproceedings{gadre2023cow,
  author    = {Gadre, Samir Yitzhak and Wortsman, Mitchell and Ilharco, Gabriel and Schmidt, Ludwig and Song, Shuran},
  title     = {{CoWs} on Pasture: Baselines and Benchmarks for Language-Driven Zero-Shot Object Navigation},
  booktitle = CVPR,
  pages     = {23171--23181},
  year      = {2023},
  url       = {https://openaccess.thecvf.com/content/CVPR2023/html/Gadre_CoWs_on_Pasture_Baselines_and_Benchmarks_for_Language-Driven_Zero-Shot_Object_CVPR_2023_paper.html},
}

@article{hou2026slider,
  author  = {Hou, Xiaolei and Pan, Zheng and Lan, Hua and Zou, Zhenghao and Chen, Yinhong and Zhu, Chenxi and Lyu, Yang and Hu, Jinwen and Zhao, Chunhui},
  title   = {{SLIDER}: Sparse History-Guided Aerial Robot Target Search Using Sliding Local Maps},
  journal = RAL,
  volume  = {11},
  number  = {9},
  pages   = {10617--10624},
  year    = {2026},
  doi     = {10.1109/LRA.2026.3715036},
  url     = {https://doi.org/10.1109/LRA.2026.3715036},
}

@article{wu2023uavrescue,
  author  = {Wu, Jiehong and Sun, Ya'nan and Li, Danyang and Shi, Jun-ling and Li, Xianwei and Gao, Lijun and Yu, Lei and Han, Guangjie and Wu, Jinsong},
  title   = {An Adaptive Conversion Speed {Q}-Learning Algorithm for Search and Rescue {UAV} Path Planning in Unknown Environments},
  journal = TVT,
  volume  = {72},
  number  = {12},
  pages   = {15391--15404},
  year    = {2023},
  doi     = {10.1109/TVT.2023.3297837},
  url     = {https://doi.org/10.1109/TVT.2023.3297837},
}

@article{tse2024uavinspection,
  author  = {Tse, Kwai-Wa and Pi, Rendong and Yang, Wenyu and Yu, Xiang and Wen, Chih-Yung},
  title   = {Advancing {UAV}-Based Inspection System: The {USSA-Net} Segmentation Approach to Crack Quantification},
  journal = TIM,
  volume  = {73},
  pages   = {1--14},
  year    = {2024},
  note    = {Art. no. 2522914},
  doi     = {10.1109/TIM.2024.3418073},
  url     = {https://doi.org/10.1109/TIM.2024.3418073},
}

@inproceedings{yamauchi1997frontier,
  author    = {Yamauchi, Brian},
  title     = {A Frontier-Based Approach for Autonomous Exploration},
  booktitle = CIRA,
  year      = {1997},
  pages     = {146--151},
  doi       = {10.1109/CIRA.1997.613851},
  url       = {https://doi.org/10.1109/CIRA.1997.613851},
}

@article{luo2024starsearcher,
  author  = {Luo, Yiming and Zhuang, Zixuan and Pan, Neng and Feng, Chen and Shen, Shaojie and Gao, Fei and Cheng, Hui and Zhou, Boyu},
  title   = {{Star-Searcher}: A Complete and Efficient Aerial System for Autonomous Target Search in Complex Unknown Environments},
  journal = RAL,
  year    = {2024},
  volume  = {9},
  number  = {5},
  pages   = {4329--4336},
  doi     = {10.1109/LRA.2024.3379840},
  url     = {https://doi.org/10.1109/LRA.2024.3379840},
}

@inproceedings{ramakrishnan2022poni,
  author    = {Ramakrishnan, Santhosh Kumar and Chaplot, Devendra Singh and Al-Halah, Ziad and Malik, Jitendra and Grauman, Kristen},
  title     = {{PONI}: Potential Functions for {ObjectGoal} Navigation With Interaction-Free Learning},
  booktitle = CVPR,
  year      = {2022},
  pages     = {18890--18900},
  url       = {https://openaccess.thecvf.com/content/CVPR2022/html/Ramakrishnan_PONI_Potential_Functions_for_ObjectGoal_Navigation_With_Interaction-Free_Learning_CVPR_2022_paper.html},
}

@inproceedings{dang2023dual,
  author    = {Dang, Ronghao and Wang, Liuyi and He, Zongtao and Su, Shuai and Tang, Jiagui and Liu, Chengju and Chen, Qijun},
  title     = {Search for or Navigate to? Dual Adaptive Thinking for Object Navigation},
  booktitle = ICCV,
  year      = {2023},
  pages     = {8250--8259},
  url       = {https://openaccess.thecvf.com/content/ICCV2023/html/Dang_Search_for_or_Navigate_to_Dual_Adaptive_Thinking_for_Object_ICCV_2023_paper.html},
}

@inproceedings{zhang2024imagine,
  author    = {Zhang, Sixian and Yu, Xinyao and Song, Xinhang and Wang, Xiaohan and Jiang, Shuqiang},
  title     = {Imagine Before Go: Self-Supervised Generative Map for Object Goal Navigation},
  booktitle = CVPR,
  year      = {2024},
  pages     = {16414--16425},
  url       = {https://openaccess.thecvf.com/content/CVPR2024/html/Zhang_Imagine_Before_Go_Self-Supervised_Generative_Map_for_Object_Goal_Navigation_CVPR_2024_paper.html},
}

@inproceedings{nie2025wmnav,
  author    = {Nie, Dujun and Guo, Xianda and Duan, Yiqun and Zhang, Ruijun and Chen, Long},
  title     = {{WMNav}: Integrating Vision-Language Models into World Models for Object Goal Navigation},
  booktitle = IROS,
  year      = {2025},
  pages     = {2392--2399},
  doi       = {10.1109/IROS60139.2025.11246684},
  url       = {https://doi.org/10.1109/IROS60139.2025.11246684},
}

@article{zhang2025anwm,
  author  = {Zhang, Weichen and Tang, Peizhi and Zeng, Xin and Man, Fanhang and Yu, Shiquan and Dai, Zichao and Zhao, Baining and Chen, Hongjin and Shang, Yu and Wu, Wei and Gao, Chen and Chen, Xinlei and Wang, Xin and Li, Yong and Zhu, Wenwu},
  title   = {Aerial World Model for Long-Horizon Visual Generation and Navigation in {3D} Space},
  journal = {arXiv preprint arXiv:2512.21887},
  year    = {2025},
  url     = {https://arxiv.org/abs/2512.21887},
}

@inproceedings{lee2025citynav,
  author    = {Lee, Jungdae and Miyanishi, Taiki and Kurita, Shuhei and Sakamoto, Koya and Azuma, Daichi and Matsuo, Yutaka and Inoue, Nakamasa},
  title     = {{CityNav}: A Large-Scale Dataset for Real-World Aerial Navigation},
  booktitle = ICCV,
  year      = {2025},
  pages     = {5912--5922},
  url       = {https://openaccess.thecvf.com/content/ICCV2025/html/Lee_CityNav_A_Large-Scale_Dataset_for_Real-World_Aerial_Navigation_ICCV_2025_paper.html},
}

@article{xu2026geonav,
  author  = {Xu, Haotian and Hu, Yue and Gao, Chen and Zhu, Zhengqiu and Zhao, Yong and Yin, Quanjun},
  title   = {{GeoNav}: Empowering {MLLMs} with Dual-Scale Geospatial Reasoning for Language-Goal Aerial Navigation},
  journal = PR,
  year    = {2026},
  volume  = {177},
  pages   = {113365},
  url     = {https://www.sciencedirect.com/science/article/pii/S0031320326003304},
}

@inproceedings{cai2025flightgpt,
  author    = {Cai, Hengxing and Dong, Jinhan and Tan, Jingjun and Deng, Jingcheng and Li, Sihang and Gao, Zhifeng and Wang, Haidong and Su, Zicheng and Sumalee, Agachai and Zhong, Renxin},
  title     = {{FlightGPT}: Towards Generalizable and Interpretable {UAV} Vision-and-Language Navigation with Vision-Language Models},
  booktitle = EMNLP,
  year      = {2025},
  pages     = {6659--6676},
  doi       = {10.18653/v1/2025.emnlp-main.338},
  url       = {https://aclanthology.org/2025.emnlp-main.338/},
}

@inproceedings{wang2023dreamwalker,
  author    = {Wang, Hanqing and Liang, Wei and Van Gool, Luc and Wang, Wenguan},
  title     = {{DREAMWALKER}: Mental Planning for Continuous Vision-Language Navigation},
  booktitle = ICCV,
  year      = {2023},
  pages     = {10873--10883},
  url       = {https://openaccess.thecvf.com/content/ICCV2023/html/Wang_DREAMWALKER_Mental_Planning_for_Continuous_Vision-Language_Navigation_ICCV_2023_paper.html},
}

@inproceedings{li2025skyvln,
  author    = {Li, Tianshun and Huai, Tianyi and Li, Zhen and Gao, Yichun and Li, Haoang and Zheng, Xinhu},
  title     = {{SkyVLN}: Vision-and-Language Navigation and {NMPC} Control for {UAVs} in Urban Environments},
  booktitle = IROS,
  year      = {2025},
  pages     = {17199--17206},
  doi       = {10.1109/IROS60139.2025.11247609},
  url       = {https://doi.org/10.1109/IROS60139.2025.11247609},
}

@inproceedings{shinn2023reflexion,
  author    = {Shinn, Noah and Cassano, Federico and Gopinath, Ashwin and Narasimhan, Karthik and Yao, Shunyu},
  title     = {Reflexion: Language Agents with Verbal Reinforcement Learning},
  booktitle = NIPS,
  year      = {2023},
  volume    = {36},
  pages     = {8634--8652},
  url       = {https://proceedings.neurips.cc/paper_files/paper/2023/hash/1b44b878bb782e6954cd888628510e90-Abstract-Conference.html},
}

@inproceedings{cao2025cognav,
  author    = {Cao, Yihan and Zhang, Jiazhao and Yu, Zhinan and Liu, Shuzhen and Qin, Zheng and Zou, Qin and Du, Bo and Xu, Kai},
  title     = {{CogNav}: Cognitive Process Modeling for Object Goal Navigation with {LLMs}},
  booktitle = ICCV,
  year      = {2025},
  pages     = {9550--9560},
  url       = {https://openaccess.thecvf.com/content/ICCV2025/html/Cao_CogNav_Cognitive_Process_Modeling_for_Object_Goal_Navigation_with_LLMs_ICCV_2025_paper.html},
}

@inproceedings{qin2025aprr,
  author    = {Qin, Liang and Wang, Min and Li, Peiwei and Zhou, Wengang and Li, Houqiang},
  title     = {Active Perception Meets Rule-Guided {RL}: A Two-Phase Approach for Precise Object Navigation in Complex Environments},
  booktitle = ICCV,
  year      = {2025},
  pages     = {7603--7612},
  doi       = {10.1109/ICCV51701.2025.00713},
  url       = {https://openaccess.thecvf.com/content/ICCV2025/html/Qin_Active_Perception_Meets_Rule-Guided_RL_A_Two-Phase_Approach_for_Precise_ICCV_2025_paper.html},
}

@inproceedings{ko2025care,
  author    = {Ko, Po-Chen and Su, Hung-Ting and Chen, CY and Yeh, Jia-Fong and Sun, Min and Hsu, Winston H.},
  title     = {Context-Aware Replanning with Pre-Explored Semantic Map for Object Navigation},
  booktitle = CORL,
  series    = PMLR,
  year      = {2025},
  volume    = {270},
  pages     = {4253--4267},
  url       = {https://proceedings.mlr.press/v270/ko25b.html},
}

@inproceedings{wang2026moge,
  author  = {Wang, Ruicheng and Xu, Sicheng and Dong, Yue and Deng, Yu and Xiang, Jianfeng and Lv, Zelong and Sun, Guangzhong and Tong, Xin and Yang, Jiaolong},
  title   = {Moge-2: Accurate monocular geometry with metric scale and sharp details},
  booktitle = NIPS,
  year    = {2025},
  url     = {https://proceedings.neurips.cc/paper_files/paper/2025/hash/336572db3e99930814d6b328d4220cb6-Abstract-Conference.html}
}

\twocolumn[{%
  \noindent\makebox[\textwidth][c]{%
    \Large\bfseries\color{mthreeblue}Supplementary Material%
  }\par
  \vspace{0.6em}
}]
\setcounter{figure}{0}
\setcounter{table}{0}
\renewcommand{\thefigure}{A\arabic{figure}}
\renewcommand{\thetable}{A\arabic{table}}
\renewcommand{\thealgocf}{A\arabic{algocf}}
\appendix

\section{Evaluation Metrics}
\label{app:metrics}

We define the metrics below while preserving each benchmark’s rules for goal proximity, trajectory sampling, and collision handling.
Let $\{\bm{x}_{i,0},\cdots,\bm{x}_{i,T_i}\}$ denote the recorded UAV positions in episode $i$, and let $E_i$ contain the measured endpoints of its high-level decisions. 
The binary indicators $C_i$ and $H_i$ denote a collision and termination with an accepted \textbf{\texttt{Stop}}, respectively. 
For any position $\bm{p}$, the goal distance is defined as:
\begin{equation}\label{eq:goal_dist}
  d_i(\bm{p}) = \min_{\bm{g}\in G_i}\|\bm{p}-\bm{g}\|_2 .
\end{equation}
Here, $G_i$ contains the annotated goal points on UAV-ON \cite{xiao2025uavon}, the target anchor on AerialDojo-200K \cite{feng2026aerialdojo}, or the annotated object position on TravelUAV \cite{wang2024openuav}. 
We use $\bm{1}\{\cdot\}$ for an indicator and $N$ for the number of scored episodes. All distances are measured in meters.

\paragraph{Success Rate (SR).}
For SimpleUAV, an episode succeeds if it terminates with an accepted \textbf{\texttt{Stop}} within the evaluation radius $\tau$ and remains collision-free:
\begin{equation}\label{eq:SR}
    \begin{aligned}
        S_i &= (1-C_i)H_i\,\bm{1}\{d_i(\bm{x}_{i,T_i})\le\tau\},\\
        \mathrm{SR} &= \frac{1}{N} \sum_{i=1}^N S_i .
    \end{aligned}
\end{equation}
We use $\tau=20$ m on UAV-ON and TravelUAV and $\tau=3$ m for the primary AerialDojo-200K evaluation.

\paragraph{Oracle Success Rate (OSR).}
OSR measures whether an episode satisfies the benchmark’s oracle proximity criterion, including cases without a successful stop. 
We \mbox{compute OSR as:}
\begingroup
\setlength{\abovedisplayskip}{\abovedisplayshortskip}
\setlength{\belowdisplayskip}{\belowdisplayshortskip}
\begin{equation}\label{eq:OSR}
  \mathrm{OSR} = \frac{1}{N} \sum_{i=1}^{N} O_i, 
\end{equation}
\endgroup
with $O_i$ defined separately for each benchmark.

For \textbf{UAV-ON}, let $B_i$ indicate exhaustion of the decision budget.
A collision-free episode earns oracle credit if it ends successfully, reaches a decision endpoint strictly within 20~m of the goal, or exhausts its budget at most 20~m \mbox{from the goal.}
\begin{equation}
    \begin{aligned}
        O_i=(1-C_i)\,\bm{1}\bigl\{&
        S_i=1\ \lor\ \exists\,\bm{p}\in E_i:\,d_i(\bm{p})<20\\
        &{}\lor\bigl(B_i=1\ \land\ d_i(\bm{x}_{i,T_i})\le20\bigr)\bigr\}.
    \end{aligned}
\end{equation}
The endpoint test excludes the starting position and intermediate controller samples. 
A collision sets $O_i=0$, even if an earlier endpoint satisfied the proximity criterion.

For \textbf{AerialDojo-200K}, oracle success considers the starting position, decision endpoints, and final position. 
Defining $A_i=E_i\cup\{\bm{x}_{i,0},\bm{x}_{i,T_i}\}$, we compute $O_i$ as:
\begin{equation}
    O_i=\bm{1}\{\exists\,\bm{p}\in A_i\ \text{with}\ d_i(\bm{p})\le3\}.
\end{equation}
A qualifying sample retains oracle credit even if the episode subsequently fails or collides.

\paragraph{Success Weighted by Path Length (SPL).}
SPL weights successful episodes by path efficiency. 
The executed path length $P_i$ sums the displacements between consecutive recorded positions, including the intermediate positions sampled by the controller:
\begingroup
\setlength{\abovedisplayskip}{\abovedisplayshortskip}
\setlength{\belowdisplayskip}{\belowdisplayshortskip}
\begin{equation}
    \begin{aligned}
        P_i &= \sum_{j=1}^{T_i}\|\bm{x}_{i,j}-\bm{x}_{i,j-1}\|_2,\\
        \mathrm{SPL} &= \frac{1}{N}\sum_{i=1}^N S_i
            \frac{L_i}{\max(L_i,P_i)} .
    \end{aligned}
\end{equation}
\endgroup
The reference length $L_i$ is the dataset-provided geodesic distance on UAV-ON and the full reference trajectory length on AerialDojo-200K. 
UAV-ON additionally assigns zero SPL contribution to episodes with zero executed path length.

\paragraph{Collision Rate (CR).}
For the UAV-ON component ablations, CR measures the fraction of all scored episodes containing a collision:
\begin{equation}
    \mathrm{CR}=\frac{1}{N}\sum_{i=1}^N C_i .
\end{equation}
The denominator includes all episode outcomes. We report SR, OSR, SPL, and CR as percentages.

\section{Implementation Details}
\label{app:implementation}

We run VLM inference with vLLM on NVIDIA RTX PRO 6000 GPUs, enabling multi-token prediction (MTP) where supported. Unless otherwise specified, the backbone is Qwen3.6-35B-A3B. All VLM weights remain frozen, with no task-specific training or fine-tuning. 
Table~\ref{tab:model_checkpoints} lists the exact Hugging Face checkpoint IDs; all listed checkpoints use FP8 quantization.

\begin{table}[!t]
\centering
\small
\renewcommand{\arraystretch}{1.3}
\caption{\textbf{Hugging Face model IDs used in our experiments.} All checkpoints use FP8 quantization.}
\label{tab:model_checkpoints}
\vspace{-0.5em}
\resizebox{1.0\linewidth}{!}{%
\begin{tabular}{ll}
  \toprule
  \rowcolor{tbhead}
  \textbf{Backbone} & \textbf{Hugging Face Checkpoint ID} \\
  \midrule
  \textbf{Qwen3.6-35B-A3B} & \texttt{Qwen/Qwen3.6-35B-A3B-FP8} \\
  \rowcolor{tbzebra} 
  \textbf{Qwen3.8-27B} & \texttt{Qwen/Qwen3.8-27B-FP8} \\
  \textbf{Qwen3.5-9B} & \texttt{RedHatAI/Qwen3.5-9B-FP8-dynamic} \\
  \rowcolor{tbzebra} 
  \textbf{Qwen3.5-4B} & \texttt{RedHatAI/Qwen3.5-4B-FP8-dynamic} \\
  \textbf{Qwen3-VL-8B-Instruct} & \texttt{Qwen/Qwen3-VL-8B-Instruct-FP8} \\
  \rowcolor{tbzebra} 
  \textbf{Qwen3-VL-4B-Instruct} & \texttt{Qwen/Qwen3-VL-4B-Instruct-FP8} \\
  \bottomrule
\end{tabular}}
\end{table}

The agent receives four RGB views, i.e., front, left, right, and downward, and retains the four most recent interaction rounds alongside a bounded text summary. Following the benchmark episode limits, the default decision budgets are 150 on UAV-ON, 90 on AerialDojo-200K Base, and 200~on~TravelUAV.

\subsection{Shared Interfaces and Controller Settings}
\label{app:shared-controller}

\paragraph{Candidate and Search Context.}
The assessment query returns at most four candidates, each represented by a tight bounding box in \textbf{\texttt{xyxy}} format with coordinates normalized to $[0,1000]$. 
Candidate identifiers are assigned anew for each observation. On UAV-ON and AerialDojo-200K, each retained interaction round contains one selected-view image, the executed action, and its outcome. 
Table~\ref{tab:controller_parameters} summarizes the controller settings across benchmarks.

\begin{table}[t]
\centering
\small
\setlength{\tabcolsep}{2pt}
\renewcommand{\arraystretch}{1.4}
\caption{\textbf{Controller settings across benchmarks.} The stopping threshold uses estimated candidate distance; the evaluation radius uses annotated goal distance.}
\label{tab:controller_parameters}
\vspace{-0.5em}
\resizebox{1.0\columnwidth}{!}{%
\begin{tabular}{lccc}
  \toprule
  \rowcolor{tbhead}
  \textbf{Parameter} & \textbf{UAV-ON} & \textbf{A200K Base} & \textbf{TravelUAV} \\
  \midrule
  \textbf{Maximum Decisions} & 150 & 90 & 200 \\
  \rowcolor{tbzebra} 
  \textbf{Maximum Movement-Segment Length} & 6\,m & 6\,m & 6\,m \\
  \textbf{Target Approach Standoff} & 2\,m & 2\,m & 2\,m \\
  \rowcolor{tbzebra} 
  \textbf{Controller Stopping Threshold $\tau_{\mathrm{stop}}$} & 10\,m & 3\,m & 20\,m \\
  \textbf{Evaluation Success Radius $\tau$} & 20\,m & 3\,m & 20\,m \\
  \rowcolor{tbzebra} 
  \textbf{Command Duration/Integration Step} & 0.20\,s & 0.20\,s & 0.20\,s \\
  \textbf{Protected Capsule Radius $r_{\mathrm{safe}}$} & 1.25\,m & 0.50\,m & 0.50\,m \\
  \rowcolor{tbzebra} 
  \textbf{Arrival Tolerance} & 0.50\,m & 0.50\,m & 0.50\,m \\
  \textbf{Per-segment Motion Timeout} & 30\,s & 60\,s & 30\,s \\
  \rowcolor{tbzebra} 
  \textbf{Text Summary Limit (Characters)} & 1,200 & 1,200 & 1,200 \\
  \bottomrule
\end{tabular}}
\end{table}

\begin{algorithm}[!t]
\caption{SimpleUAV search and execution.}
\label{alg:simpleuav}
\small
\addtolength{\baselineskip}{0.1em}
\SetKwComment{Comment}{$\triangleright$\ }{}
\SetKwProg{Fn}{Function}{:}{}
\KwIn{Goal $g$, decision budget $T_{\max}$, history length $K_{\mathrm{hist}}$}
\KwOut{Termination status and measured trajectory $\mathcal{P}$}
\Comment{\textbf{Search loop}}
$\mathcal{P} \gets [\bm{x}_0]$ using the measured start position\;
$(\mathcal{H}_1,m_0,f_0) \gets (\emptyset,\emptyset,\emptyset)$\;
\For{$t \in \{1,\cdots,T_{\max}\}$}{
    \If{a collision is detected}{
        \Return $(\textsc{Collision},\mathcal{P})$\;
    }
    $\mathcal{O}_t \gets$ \texttt{ObserveFourViews}()\;
    $\mathcal{B}_t \gets$ \texttt{AssessCandidates}$(g,\mathcal{I}_t)$\;
    Ground $\mathcal{B}_t$ with Eq.~\ref{eq:candidate_grounding} to form $\mathcal{C}_t$\;
    $\mathcal{K}_t \gets (\bm{x}_t,\mathcal{C}_t,\mathcal{H}_t,m_{t-1},f_{t-1})$\;
    $\mathcal{A}_t^{\mathrm{valid}} \gets$ \texttt{AllowedActions}$(\mathcal{O}_t,\mathcal{K}_t)$\;
    $(a_t,m_t) \gets$ \texttt{SelectAction}$(g,\mathcal{I}_t,\mathcal{K}_t,\mathcal{A}_t^{\mathrm{valid}})$\;
    $(\bm{x}_t^{\mathrm{end}},f_t) \gets$ \texttt{Exec}$(a_t,\mathcal{O}_t,\mathcal{C}_t)$\;
    Append measured samples through $\bm{x}_t^{\mathrm{end}}$ to $\mathcal{P}$\;
    \If{$f_t$ signals an accepted stop or collision}{
        \Return $(f_t.\mathrm{status},\mathcal{P})$\;
    }
    Let $v_t$ be the view selected by $a_t$\;
    $\mathcal{H}_{t+1} \gets$ \texttt{KeepLast}$(\mathcal{H}_t,(I_t^{v_t},a_t,f_t),K_{\mathrm{hist}})$\;
}
\Return $(\textsc{DecisionLimit},\mathcal{P})$\;

\BlankLine
\Comment{\textbf{Action execution}}
\Fn{\textup{\texttt{Exec}}$(a_t,\mathcal{O}_t,\mathcal{C}_t)$}{
    \If{$a_t$ is \textup{\texttt{Stop}}}{
        $\bm{x}_t^{\mathrm{end}} \gets$ \texttt{MeasurePosition}()\;
        \Return $(\bm{x}_t^{\mathrm{end}},\textup{\texttt{CheckStop}}(a_t,\mathcal{C}_t))$\;
    }
    $\bm{w}_t \gets$ \texttt{BuildWaypoint}$(a_t,\mathcal{O}_t,\mathcal{C}_t)$\;
    \While{within motion limits and $\bm{w}_t$ is unreached}{
        Read pose and add fresh depth to $\mathcal{D}_t^{\mathrm{exec}}$\;
        Form the next short segment toward $\bm{w}_t$\;
        \textbf{break} if the depth check blocks the segment\;
        Execute the command and log positions\;
        \textbf{break} on collision\;
    }
    Measure $\bm{x}_t^{\mathrm{end}}$ and record execution feedback $f_t$\;
    \Return $(\bm{x}_t^{\mathrm{end}},f_t)$\;
}
\end{algorithm}

\begin{table*}[!t]
\centering
\small
\renewcommand{\arraystretch}{1.18}
\caption{\textbf{Components of aerial object-search agents.} A cross ($\times$) denotes an absent component in the reported method. SearchWorld also refines its cloned policy using spatial values from imagined rollouts.}
\label{tab:search_agent_components}
\vspace{-0.5em}
\resizebox{0.99\textwidth}{!}{%
\begin{tabular}{lcccc}
  \toprule
  \rowcolor{tbhead}
  \textbf{Method} & \textbf{Explicit persistent spatial representation} & \textbf{Learned world model} & \textbf{Demonstration training} & \textbf{RL training} \\
  \midrule
  \textbf{APEX~\cite{zhang2026apex}} & 3D spatio-semantic maps & $\times$ & $\times$ & PPO \\
  \textbf{OctMem-Agent~\cite{zhou2026octmem}} & Adaptive octree memory & $\times$ & OpenVLA fine-tuning & $\times$ \\
  \textbf{AeroBelief~\cite{xiao2026aerobelief}} & Semantic belief and coverage maps & $\times$ & $\times$ & $\times$ \\
  \textbf{SearchWorld~\cite{ji2026searchworld}} & BEV exploration and obstacle maps & Recurrent state-space model & Behavior cloning & $\times$ \\
  \rowcolor{tbzebra}
  \textbf{SimpleUAV (Ours)} & $\times$ & $\times$ & $\times$ & $\times$ \\
  \bottomrule
\end{tabular}}
\vspace{-1.5em}
\end{table*}

\paragraph{Motion and Stopping.}
The controller stopping threshold $\tau_{\mathrm{stop}}$ applies to the estimated candidate distance, whereas the evaluation radius $\tau$ applies to the distance from the UAV to the annotated goal (Appendix~\ref{app:metrics}). 
The \textbf{\texttt{CheckStop}} routine accepts termination only for a confirmed, validly localized candidate whose estimated distance is at most $\tau_{\mathrm{stop}}$.

During motion, the protected capsule radius $r_{\mathrm{safe}}$ accounts for the UAV body radius and an additional clearance margin. 
On UAV-ON, an ascent guard permits at most two upward exploration actions per horizontal anchor and resets after 3 m of horizontal displacement. 
Descending motion requires depth coverage. 
Explicitly selected ascent is permitted without upward depth coverage because the UAV has \mbox{no upward-facing camera.}

When motion stalls, the controller uses current depth and execution history to identify short recovery movements. 
On UAV-ON, if missing depth coverage or a nearby obstacle interrupts a descending \textbf{\texttt{Approach}}, the controller may retry horizontally from the measured endpoint using the remaining travel budget. 
Algorithm~\ref{alg:simpleuav} summarizes the search loop and action execution.

\subsection{Cross-Benchmark Adaptation}
\label{app:cross_benchmark_adaptation}

\paragraph{AerialDojo-200K.}
On the SemanticOGS ID Base split, SimpleUAV reuses the UAV-ON prompts and decision loop. 
We construct the task instruction from the released target name, category, landmark, description, and direction. MoGe2 supplies depth for candidate localization and motion checks; Appendix~\ref{app:aerialdojo_depth} describes its configuration and the simulator diagnostics motivating this choice. 

For pixel $\bm{u}=(u_x,u_y)$, we convert predicted optical-axis depth $z_t^v(\bm{u})$ to radial distance using the focal length $f_{\mathrm{cam}}$ and principal point $(c_x,c_y)$:
\begin{equation}
    D_t^v(\bm{u})=z_t^v(\bm{u})
    \sqrt{1+\frac{(u_x-c_x)^2+(u_y-c_y)^2}{f_{\mathrm{cam}}^2}}.
    \label{eq:aerialdojo_radial_depth}
\end{equation}

\paragraph{TravelUAV.}
Both VLM queries receive the full public instruction. Candidate assessment checks whether a physical object is visible and whether its class and attributes match the instruction. 
An accepted \textbf{\texttt{Stop}} requires a confirmed, depth-localized target within the estimated 20 m stopping threshold, which the controller rechecks before termination. 
The NavFoM, TravelUAV, and AirForesight results in Table~\ref{tab:traveluav_main} are taken directly from published evaluations \mbox{under L1 assistance.}

\paragraph{Output Validation.}
Programmatically constructed JSON schemas check action types, candidate identifiers, bounding boxes, and benchmark-specific constraints. 
If validation fails, the implementation requests a corrected JSON response using the current observation and validation error.

\subsection{Baseline Reproduction and Comparison}
\label{app:baseline_reproduction}

\paragraph{AerialDojo Framework on UAV-ON.}
We evaluate AerialDojo's reference framework \cite{feng2026aerialdojo} on UAV-ON using Qwen3-VL-8B served through vLLM with an FP8 checkpoint, replacing the official example's Transformers/BF16 inference configuration. 
This setup supports four concurrent, independent UAV-ON simulator instances to \mbox{increase evaluation throughput.}

\paragraph{OctMem-Agent on AerialDojo-200K.}
We reuse the UAV-ON-trained OctMem-Agent checkpoint \cite{zhou2026octmem} without additional training. 
We retain its action categories but execute all forward, leftward, rightward, upward, and downward translations as 1 m steps. 
This mapping replaces the original 3 m, 6 m, and 9 m translations, including both longer forward actions. 
The smaller increments are intended to support finer positioning in AerialDojo-200K's compact environments and reduce the risk of overshooting the 3 m success region. 
Left and right turns change yaw by $15^\circ$, and \textbf{\texttt{Stop}} remains available. 
OctMem-Agent uses the same MoGe2 depth source as SimpleUAV (Appendix~\ref{app:aerialdojo_depth}).

\paragraph{Components of Prior Search Agents.}
Table~\ref{tab:search_agent_components} compares the spatial representations and training requirements of four prior aerial search agents. 
All accumulate spatial information across decisions: AeroBelief supplies belief and coverage maps to a prompted VLM planner; APEX and OctMem-Agent condition task-trained policies on spatial representations; and SearchWorld uses spatial memory to ground imagined rollouts.

SimpleUAV instead maintains search context through recent selected views, execution outcomes, and a bounded text summary. 
Its frozen VLM selects actions using this context, while current depth supports candidate localization and motion checks over short segments. 
Without persistent spatial maps or task-specific training, SimpleUAV achieves 41.61\% SR on UAV-ON, compared with 23.80\% for SearchWorld, the strongest of these four agents by SR (Table~\ref{tab:uavon_main}).

\section[Early Stopping in the AerialDojo Reference Framework]{Early Stopping in the AerialDojo\\Reference Framework}
\label{app:early_stopping}

We examine when AerialDojo's reference framework issues \textbf{\texttt{Stop}} in the original AerialDojo-200K environment using MoGe2-estimated depth, with Qwen3-VL-4B and Qwen3-VL-8B as backbones. 
Table~\ref{tab:aerialdojo_early_stopping} reports cumulative stopping frequencies across all evaluated episodes, including those that terminate without movement.

With Qwen3-VL-4B, 7.11\% of episodes stop without moving and 58.19\% stop within five executed steps. 
The corresponding proportions for Qwen3-VL-8B are 1.27\% and 47.45\%. 
These frequent early terminations curtail exploration and limit opportunities to collect additional evidence for target verification.

\begin{table}[!t]
\centering
\small
\setlength{\tabcolsep}{4pt}
\renewcommand{\arraystretch}{1.3}
\caption{\textbf{Early stopping in the AerialDojo reference framework.} Each backbone is evaluated in the original AerialDojo-200K environment with MoGe-estimated depth.}
\label{tab:aerialdojo_early_stopping}
\vspace{-0.5em}
\begin{tabular}{lcc}
  \toprule
  \rowcolor{tbhead} \textbf{\texttt{Stop} Timing} & \textbf{Qwen3-VL-4B} & \textbf{Qwen3-VL-8B} \\
  \midrule
  \textbf{Without Movement} & 7.11\% (359) & 1.27\% (64) \\
  \rowcolor{tbzebra} \textbf{Within 3 Steps} & 45.31\% (2,287) & 32.26\% (1,628) \\
  \textbf{Within 5 Steps} & 58.19\% (2,937) & 47.45\% (2,395) \\
  \bottomrule
\end{tabular}
\end{table}

\begingroup
\titlespacing*{\section}{0em}{0.65em}{0.4em}
\section{Depth Setup and Simulator Diagnostics}
\label{app:aerialdojo_depth}

\paragraph{Shared Depth Input for Our Evaluations.}\leavevmode\newline
On AerialDojo-200K, SimpleUAV and all baselines\newline
reproduced by us use RGB-based depth predictions from MoGe2-Large-Normal \cite{wang2026moge}. 
We adopt this depth source because simulator-provided depth in the packaged environments used for our experiments was inconsistent with known scene geometry. 
This shared configuration applies only to our reproduced results; benchmark-reported results in Table~\ref{tab:aerialdojo_main} retain their original settings. 
Navigation metrics follow the benchmark-specific protocols in Appendix~\ref{app:metrics}.

\endgroup

\paragraph{Controlled Geometric Validation.}
We assessed simulator depth in a separate diagnostic setup by placing a virtual planar surface at a known distance in front of the camera. 
The surface was used only for calibration. We recorded the actual camera and object poses and independently verified the surface distance using nine physical distance-sensor rays. 
A sample was accepted only when all nine measurements agreed with the expected geometry within 1 cm.

Table~\ref{tab:aerialdojo_depth_check} presents representative measurements with the surface 2.5 m from the camera. Each value is the median over the central $6\times6$ pixels of the raw depth response, before navigation or collision-check processing. Both depth channels deviate from the known distance, with particularly large errors in Forest. In a broader scan, 41 of the 42 scene packages completed geometric calibration, and all 41 failed at least one depth check. The remaining scene did not complete calibration and was excluded from this count.

\begin{table}[t]
\centering
\small
\setlength{\tabcolsep}{6pt}
\renewcommand{\arraystretch}{1.3}
\caption{\textbf{Simulator depth versus known geometry in AerialDojo-200K.} A calibration surface is 2.50\,m in front of the camera. All values are in meters. The 65.54\,m value reaches the transport saturation limit.}
\label{tab:aerialdojo_depth_check}
\vspace{-0.5em}
\begin{tabular}{lccc}
  \toprule
  \rowcolor{tbhead}
  \textbf{Scene} & \textbf{Known} & \textbf{Perspective} & \textbf{Planar} \\
  \midrule
  \textbf{\texttt{N\_Island\_0}} & 2.50 & 3.41 & 3.42 \\
  \rowcolor{tbzebra} \textbf{\texttt{N\_Forest\_0}} & 2.50 & 65.54 & 57.84 \\
  \bottomrule
\end{tabular}
\end{table}

\paragraph{Environment Version and Diagnostic Settings.}
These observations apply to AerialDojo-200K dataset revision \texttt{876e94f0} with ProjectAirSim Python client 1.1.0. 
The controlled checks used a diagnostic configuration with reduced Frame scale and disabled streaming. 
Supplementary checks retaining the official camera extrinsics and capture settings also reproduced incorrect depth values.

\paragraph{MoGe2 Inference Settings.}
We process each $640\times640$ RGB view independently at resolution level 9. Camera intrinsics convert predicted optical-axis depth to radial metric distance using Equation~\ref{eq:aerialdojo_radial_depth}. 
In the deployed SimpleUAV pipeline, mask-rejected, non-finite, and non-positive depth values are replaced with 60 m after radial conversion.

\begingroup
\titlespacing*{\paragraph}{0em}{0.4em}{0.75em}
\section{More Case Studies}
\label{app:more_case_studies}

We present successful and failed episodes across the three benchmarks to illustrate exploration, candidate verification, and recovery after blocked motion. 
Search context guides subsequent actions, while candidate assessment uses the current views at each decision. 
Reported target distances are measured against the benchmark annotations.

\begin{figure*}[!t]
  \centering
  \begin{subfigure}[t]{0.49\textwidth}
    \centering
    \includegraphics[width=\linewidth]{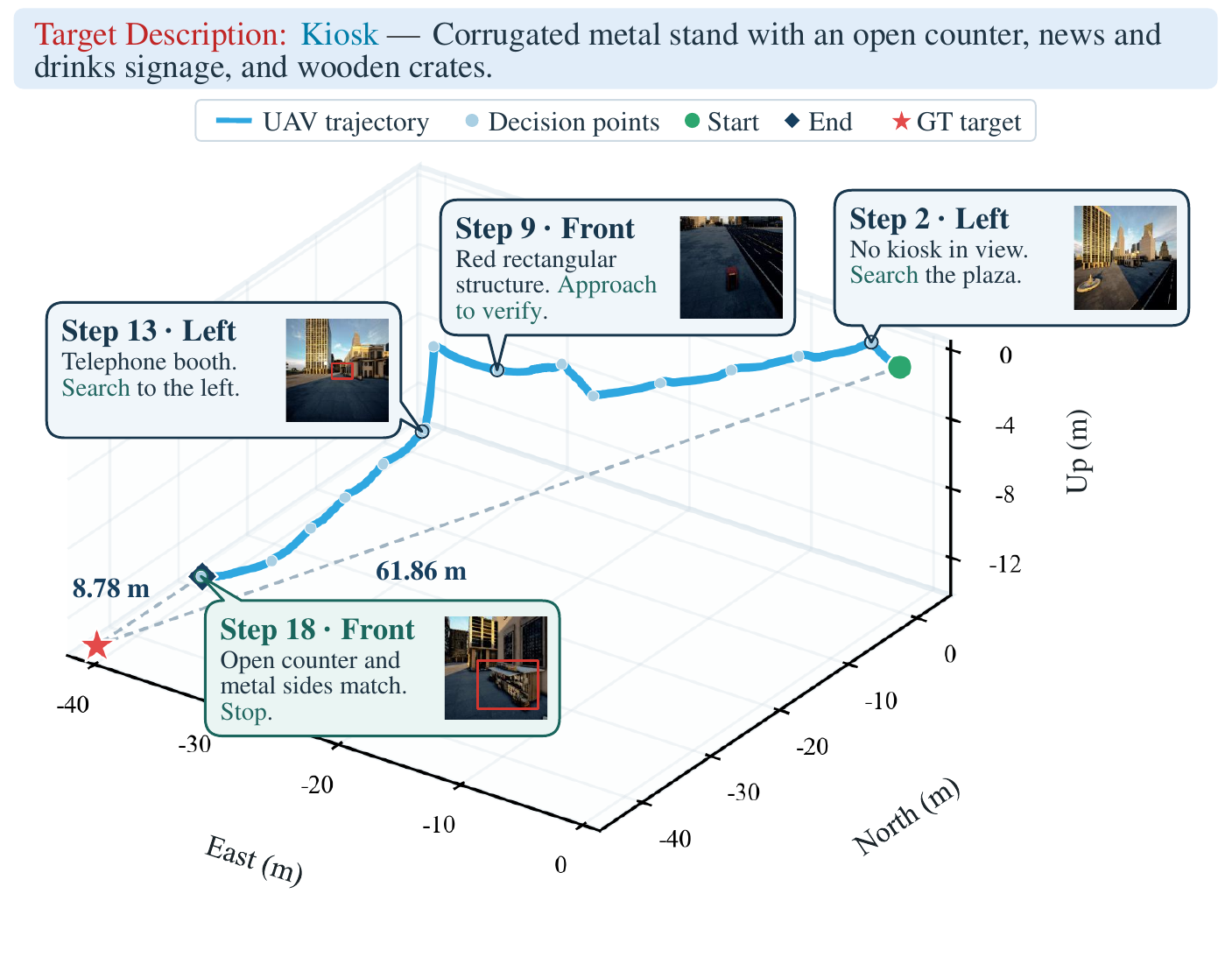}
    \vspace{-2.0em}
    \caption{\textbf{Kiosk search.} Closer inspection identifies an initial candidate as a telephone booth. The UAV resumes searching and confirms the kiosk from its open counter and metal sides.}
    \label{fig:case_study_kiosk}
  \end{subfigure}\hfill
  \begin{subfigure}[t]{0.49\textwidth}
    \centering
    \includegraphics[width=\linewidth]{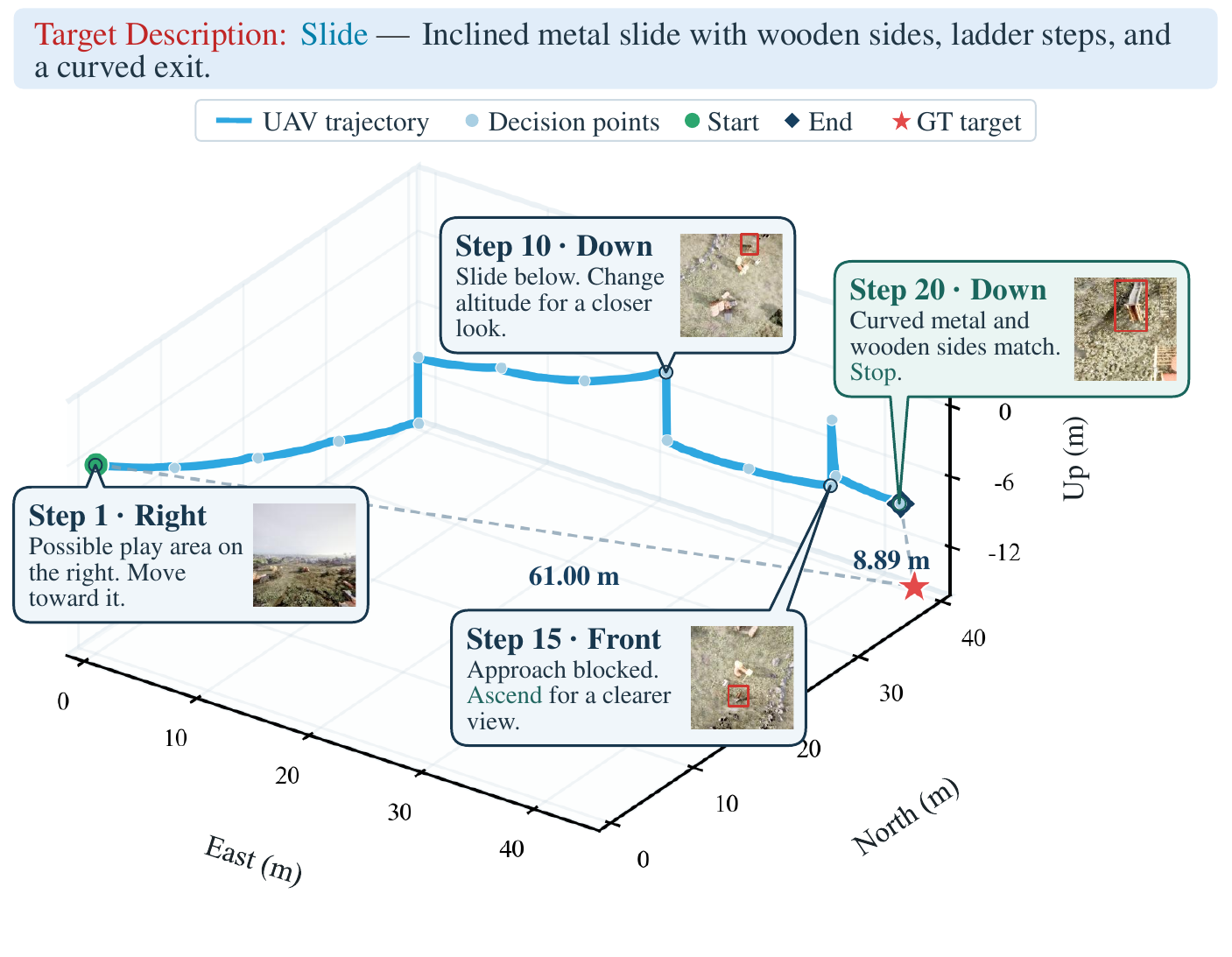}
    \vspace{-2.0em}
    \caption{\textbf{Slide search.} After a blocked approach, the UAV ascends. The new downward view reveals the curved metal surface and wooden sides, supporting slide verification.}
    \label{fig:case_study_slide}
  \end{subfigure}
  \\
  \vspace{1.0em}
  \begin{subfigure}[t]{0.49\textwidth}
    \centering
    \includegraphics[width=\linewidth]{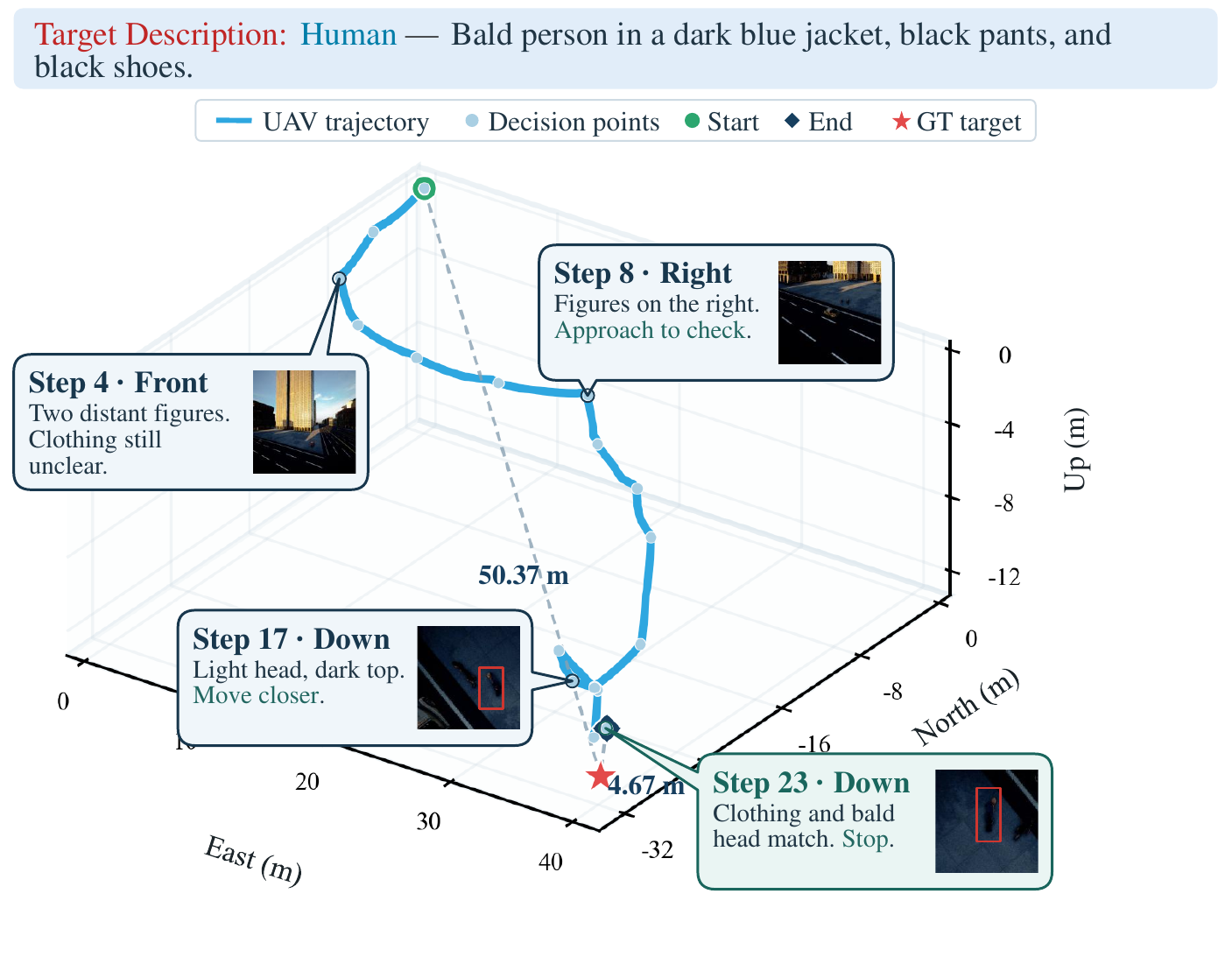}
    \vspace{-2.0em}
    \caption{\textbf{Person search.} Distant figures guide the approach. Closer downward views reveal the bald head and matching clothing needed for target confirmation.}
    \label{fig:case_study_person}
  \end{subfigure}\hfill
  \begin{subfigure}[t]{0.49\textwidth}
    \centering
    \includegraphics[width=\linewidth]{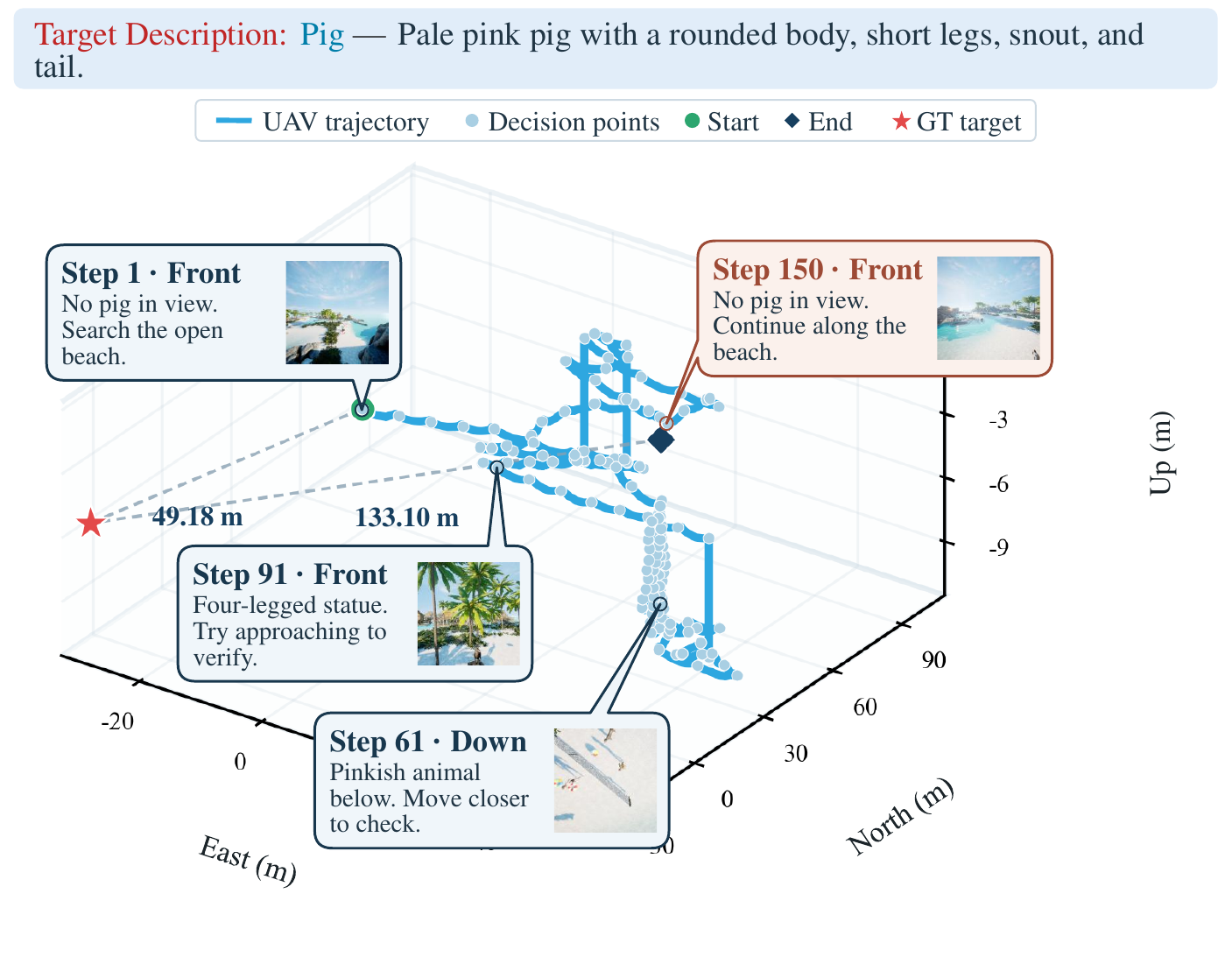}
    \vspace{-2.0em}
    \caption{\textbf{Pig search.} The UAV inspects a pinkish, animal-like object and a four-legged statue while exploring the beach, but exhausts its decision budget without confirming the target.}
    \label{fig:case_study_pig}
  \end{subfigure}
  \vspace{-0.5em}
  \caption{\textbf{Case studies on UAV-ON.} Each panel shows the executed trajectory, selected observations and decisions, and \mbox{annotated target location.}}
  \label{fig:uavon_case_studies}
  \vspace{-1.0em}
\end{figure*}

\begin{figure*}[!t]
  \centering
  \begin{subfigure}[t]{0.49\textwidth}
    \centering
    \includegraphics[width=\linewidth]{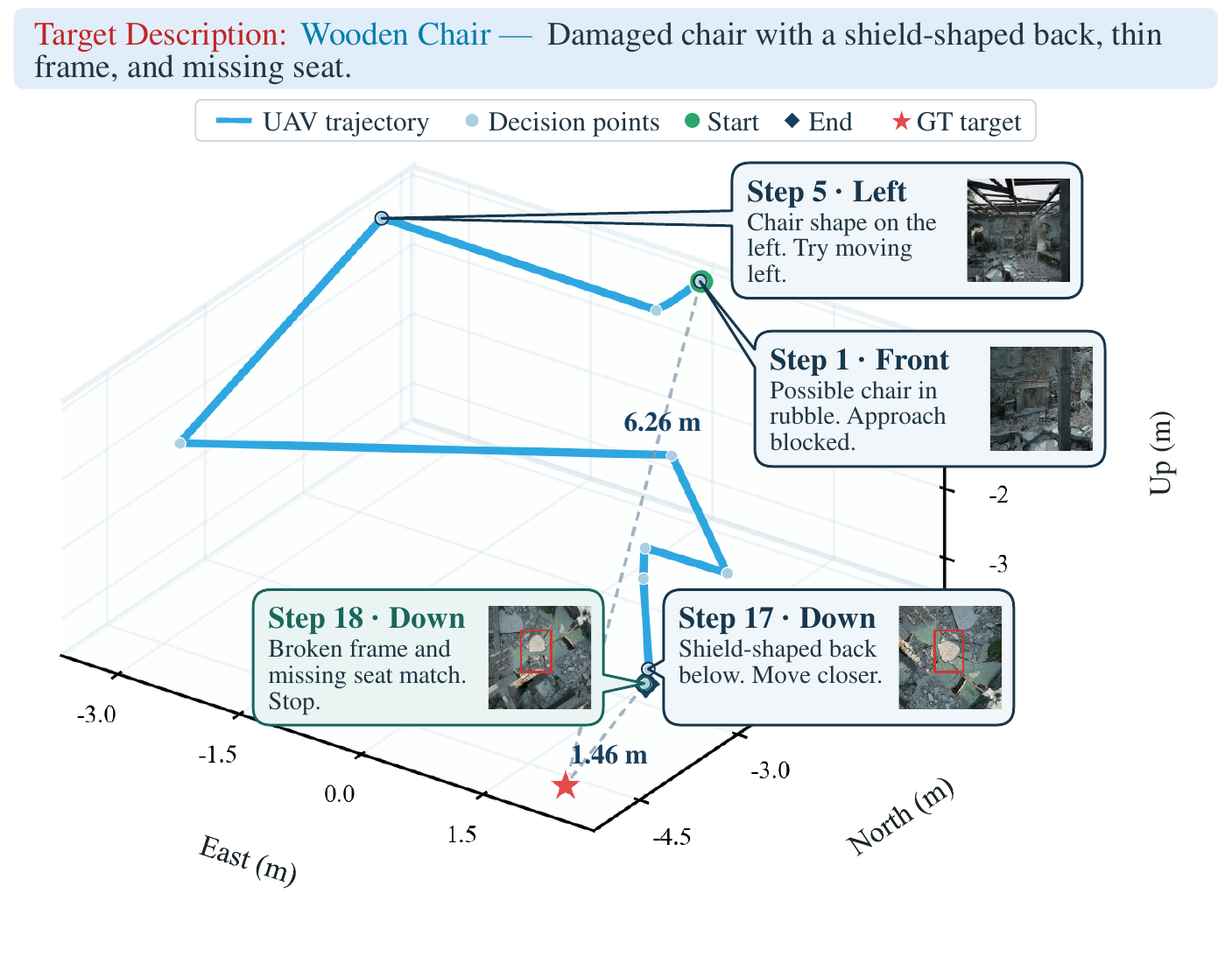}
    \vspace{-2.0em}
    \caption{\textbf{Wooden chair search.} After a blocked approach, the UAV moves left and inspects the chair from above. Its shield-shaped back, broken frame, and missing seat support confirmation and a stop 1.46 m from the target.}
    \label{fig:aerialdojo_case_chair}
  \end{subfigure}\hfill
  \begin{subfigure}[t]{0.49\textwidth}
    \centering
    \includegraphics[width=\linewidth]{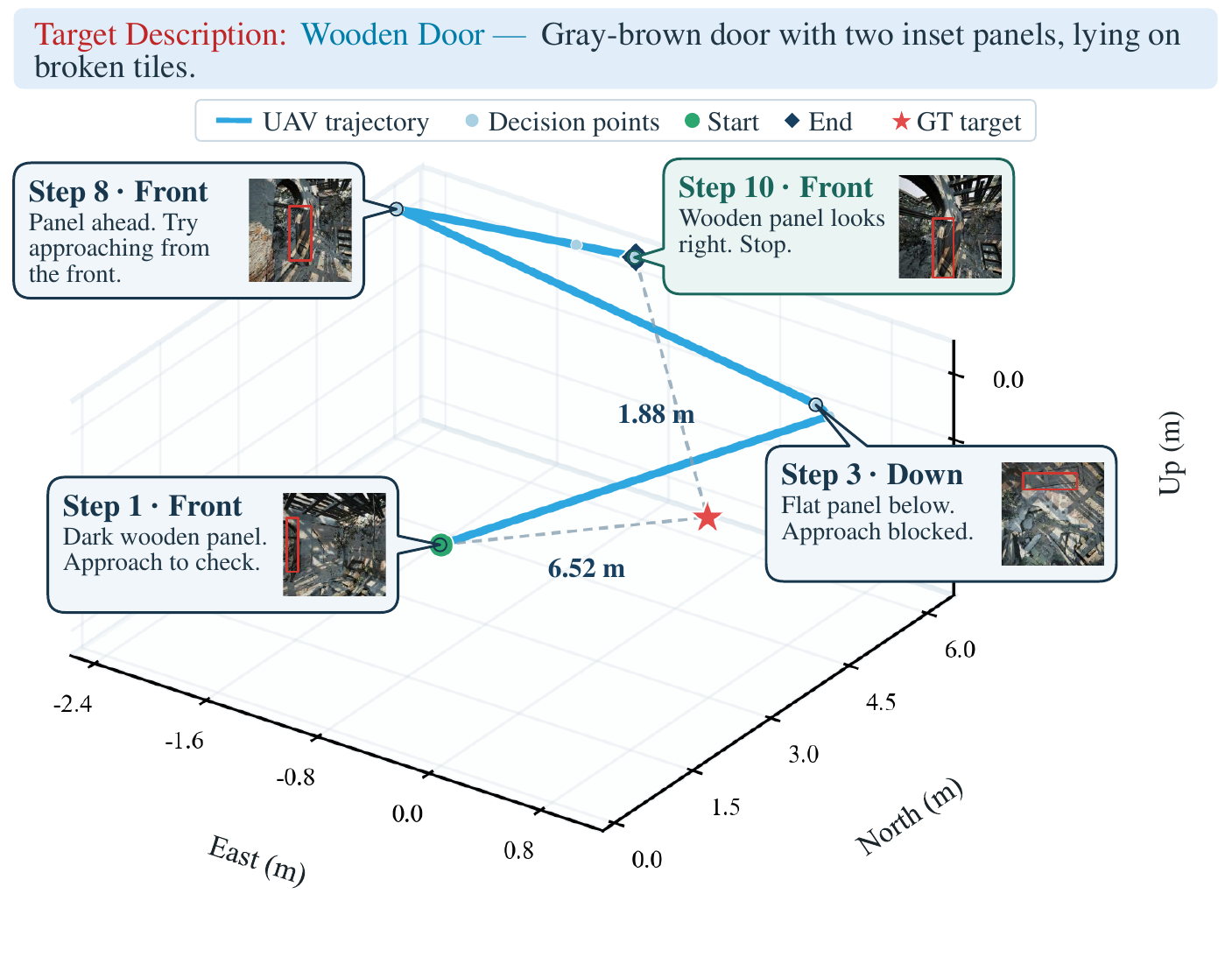}
    \vspace{-2.0em}
    \caption{\textbf{Wooden door search.} A downward view reveals the door lying on broken tiles. Following a blocked approach, the UAV repositions, approaches from the front, and stops 1.88 m from the target.}
    \label{fig:aerialdojo_case_door}
  \end{subfigure}
  \\
  \vspace{1.0em}
  \begin{subfigure}[t]{0.49\textwidth}
    \centering
    \includegraphics[width=\linewidth]{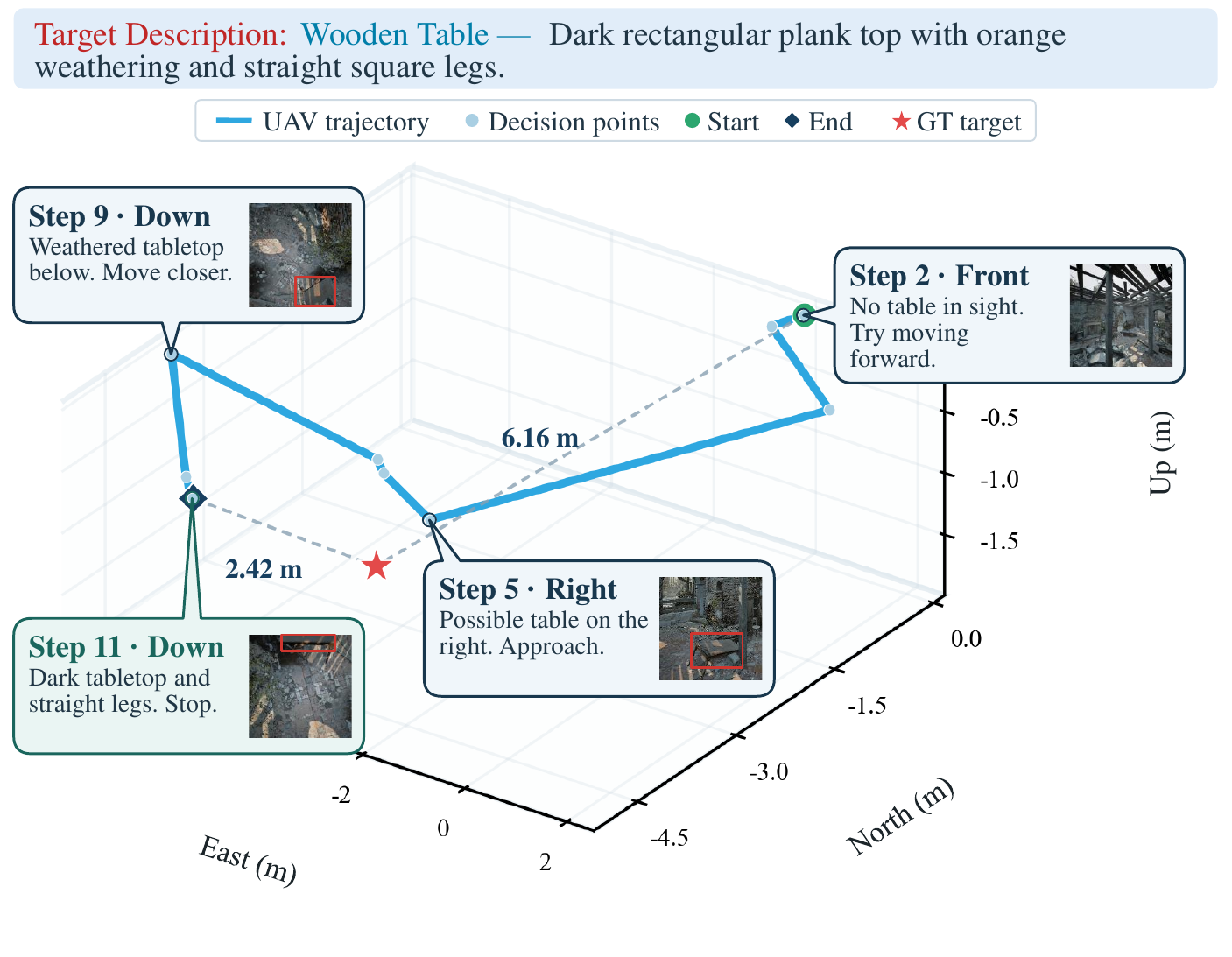}
    \vspace{-2.0em}
    \caption{\textbf{Wooden table search.} A candidate in the right view prompts an approach. Downward views reveal the dark, weathered top and straight legs, supporting confirmation and a stop \mbox{2.42 m from the target.}}
    \label{fig:aerialdojo_case_table}
  \end{subfigure}\hfill
  \begin{subfigure}[t]{0.49\textwidth}
    \centering
    \includegraphics[width=\linewidth]{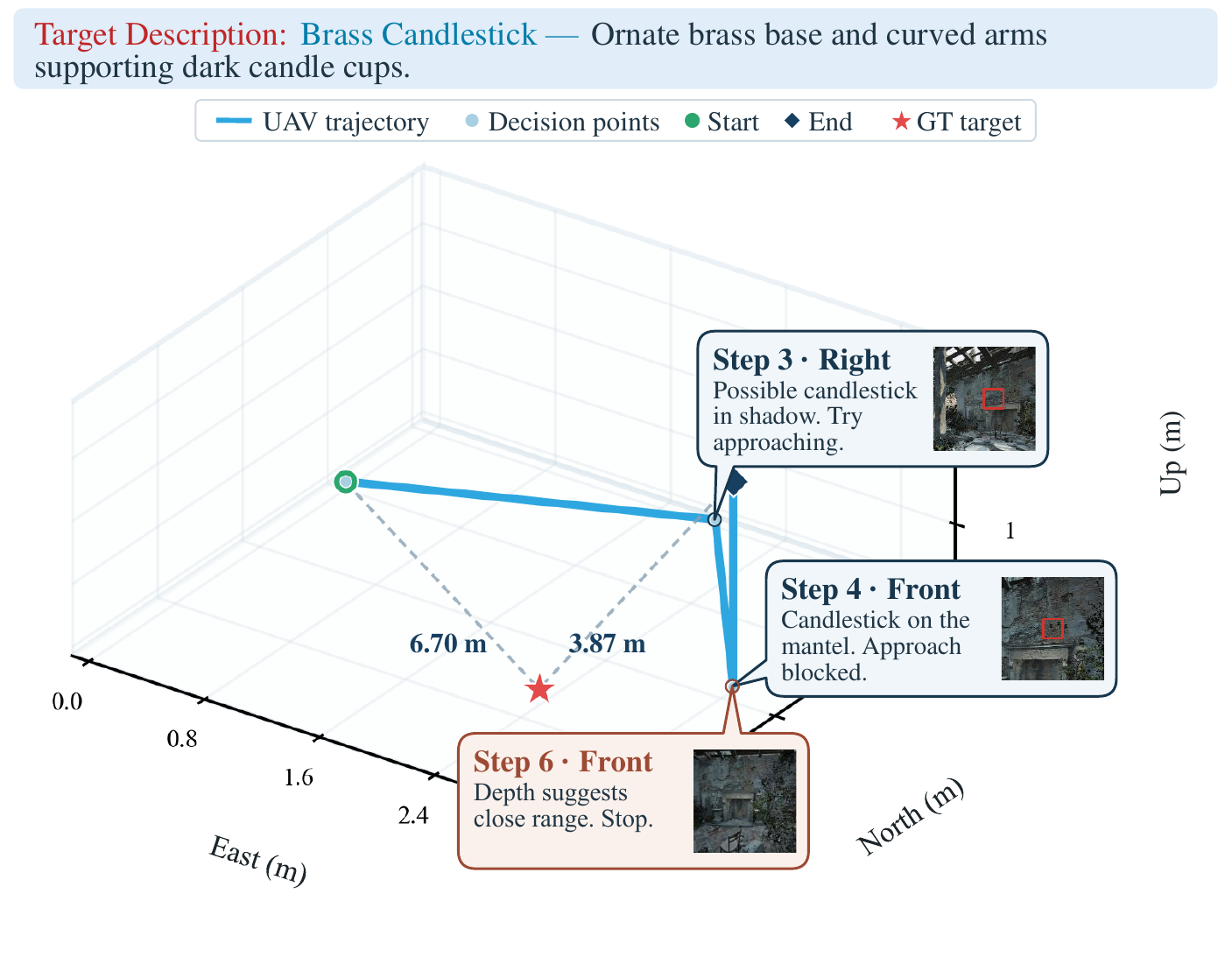}
    \vspace{-2.0em}
    \caption{\textbf{Brass candlestick search.} The UAV correctly identifies the candlestick on the mantel, but an erroneous depth estimate causes a premature stop 3.87 m from the target.}
    \label{fig:aerialdojo_case_candlestick}
  \end{subfigure}
  \vspace{-0.5em}
  \caption{\textbf{Case studies on AerialDojo-200K.} Each panel shows the executed trajectory, selected observations and decisions, and \mbox{annotated target location.}}
  \label{fig:aerialdojo_case_studies}
\end{figure*}

\begin{figure*}[!t]
  \centering
  \begin{subfigure}[t]{0.49\textwidth}
    \centering
    \includegraphics[width=0.94\linewidth]{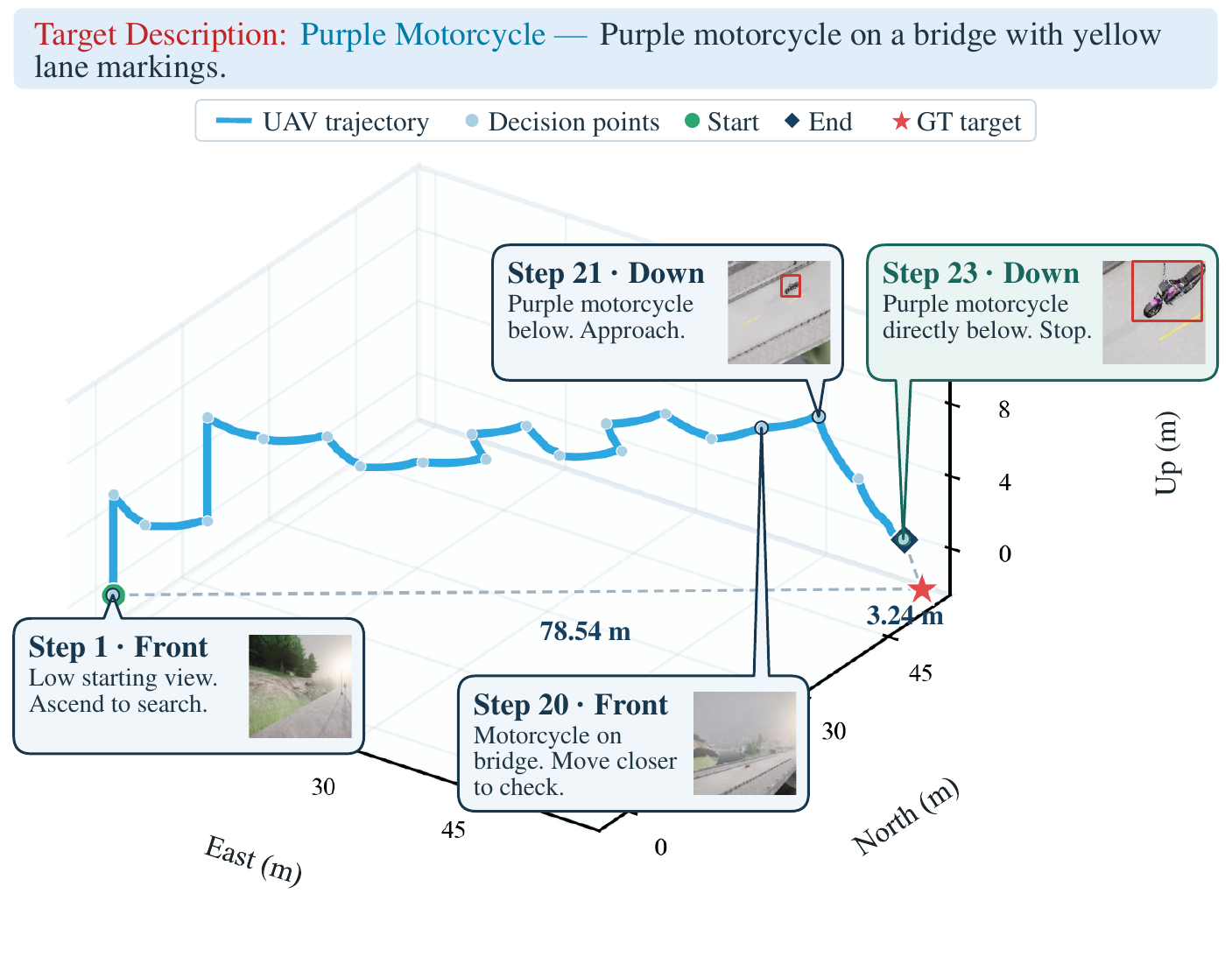}
    \vspace{-2.0em}
    \caption{\textbf{Purple motorcycle search.} The bridge and yellow lane markings guide exploration. Downward views support motorcycle confirmation as the UAV reduces its target distance from~78.54\,m to~3.24\,m.}
    \label{fig:traveluav_case_motorcycle}
  \end{subfigure}\hfill
  \begin{subfigure}[t]{0.49\textwidth}
    \centering
    \includegraphics[width=0.94\linewidth]{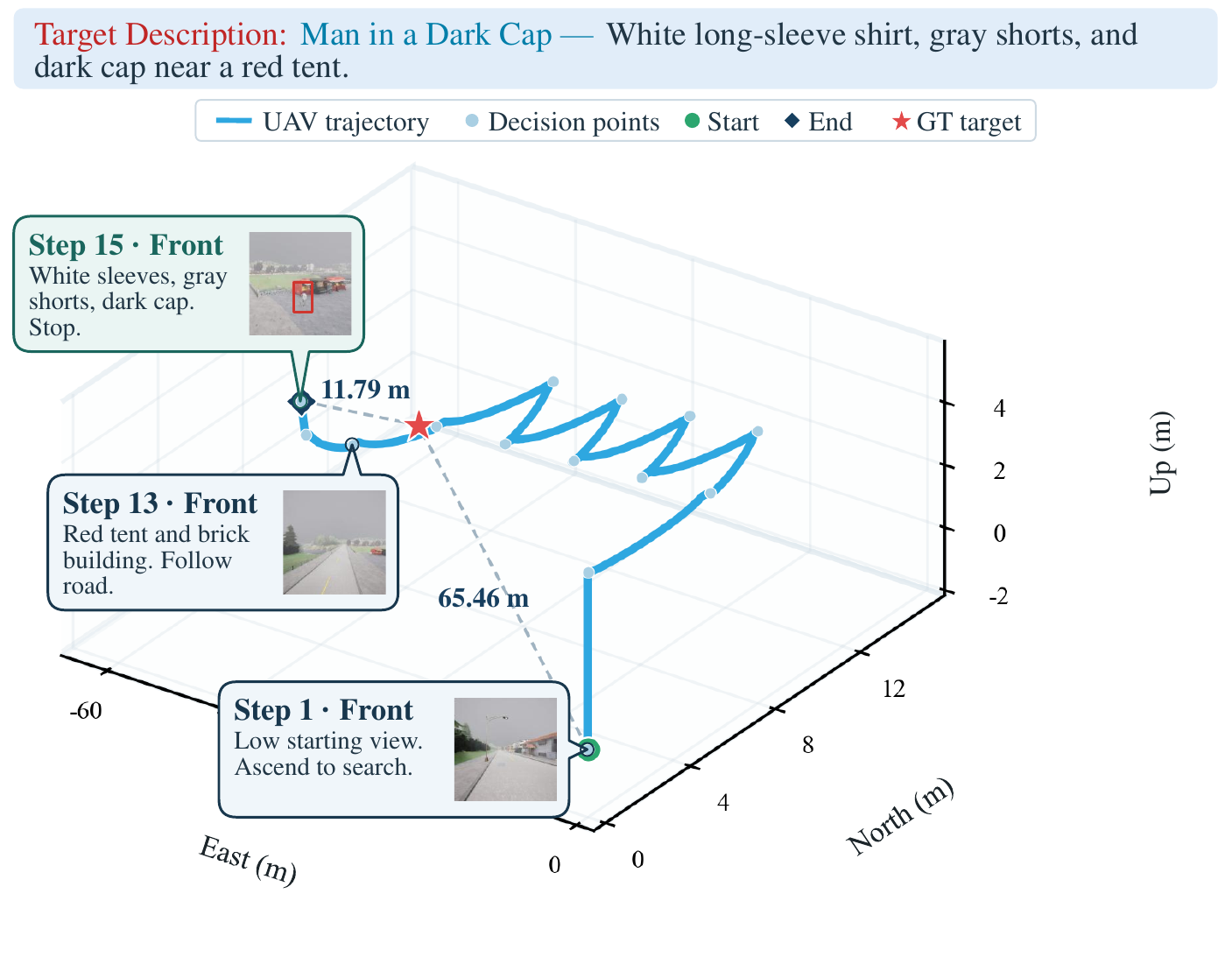}
    \vspace{-2.0em}
    \caption{\textbf{Search for a man in a dark cap.} A red tent guides the UAV along the road. White sleeves, gray shorts, and a dark cap support identification and a stop 11.79 m from the target.}
    \label{fig:traveluav_case_man}
  \end{subfigure}
  \\
  \vspace{1.0em}
  \begin{subfigure}[t]{0.49\textwidth}
    \centering
    \includegraphics[width=0.94\linewidth]{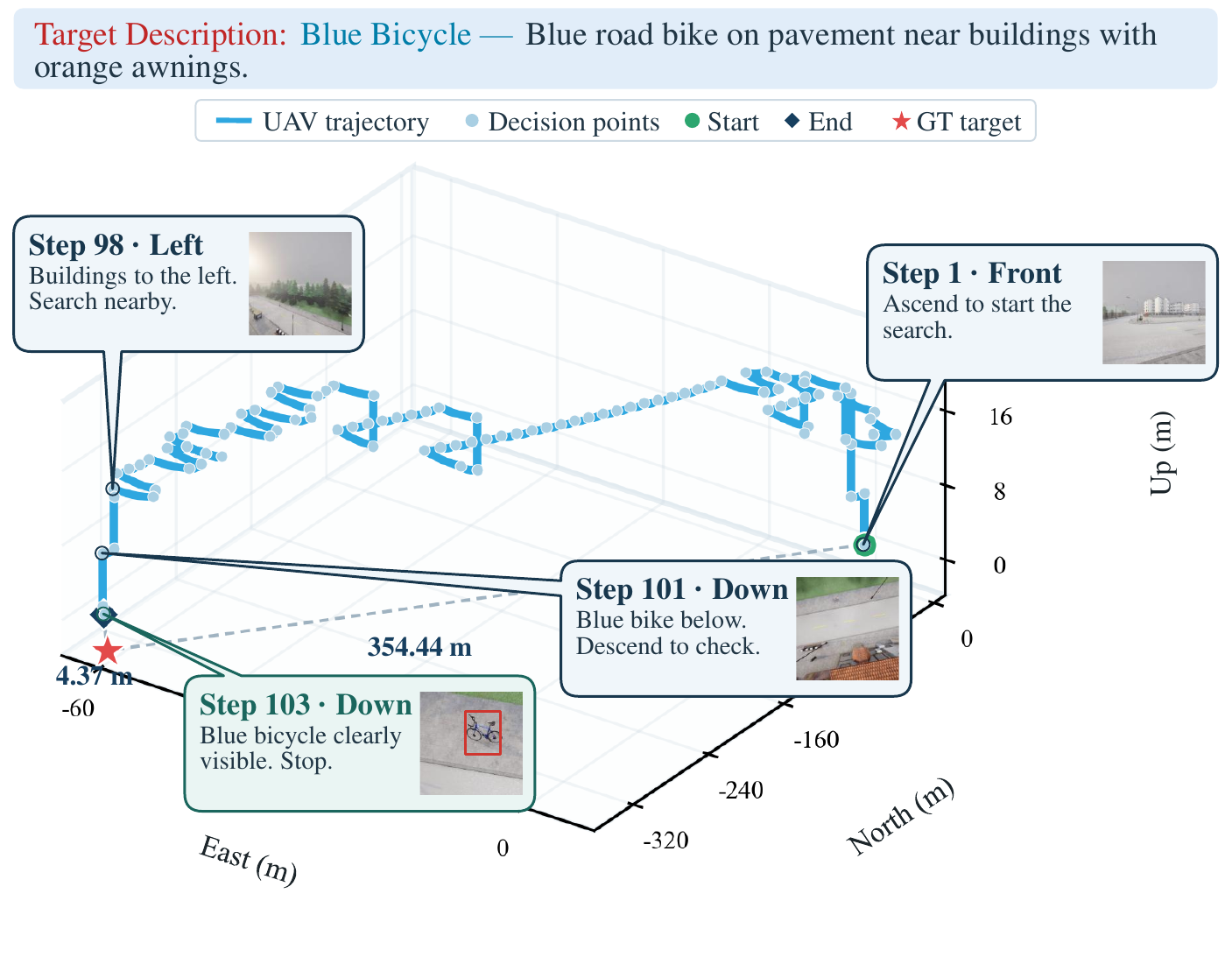}
    \vspace{-2.0em}
    \caption{\textbf{Blue bicycle search.} Starting 354.44 m away, the UAV searches near buildings to its left and verifies the bicycle from a closer downward view, stopping 4.37 m from the target.}
    \label{fig:traveluav_case_bicycle}
  \end{subfigure}\hfill
  \begin{subfigure}[t]{0.49\textwidth}
    \centering
    \includegraphics[width=0.94\linewidth]{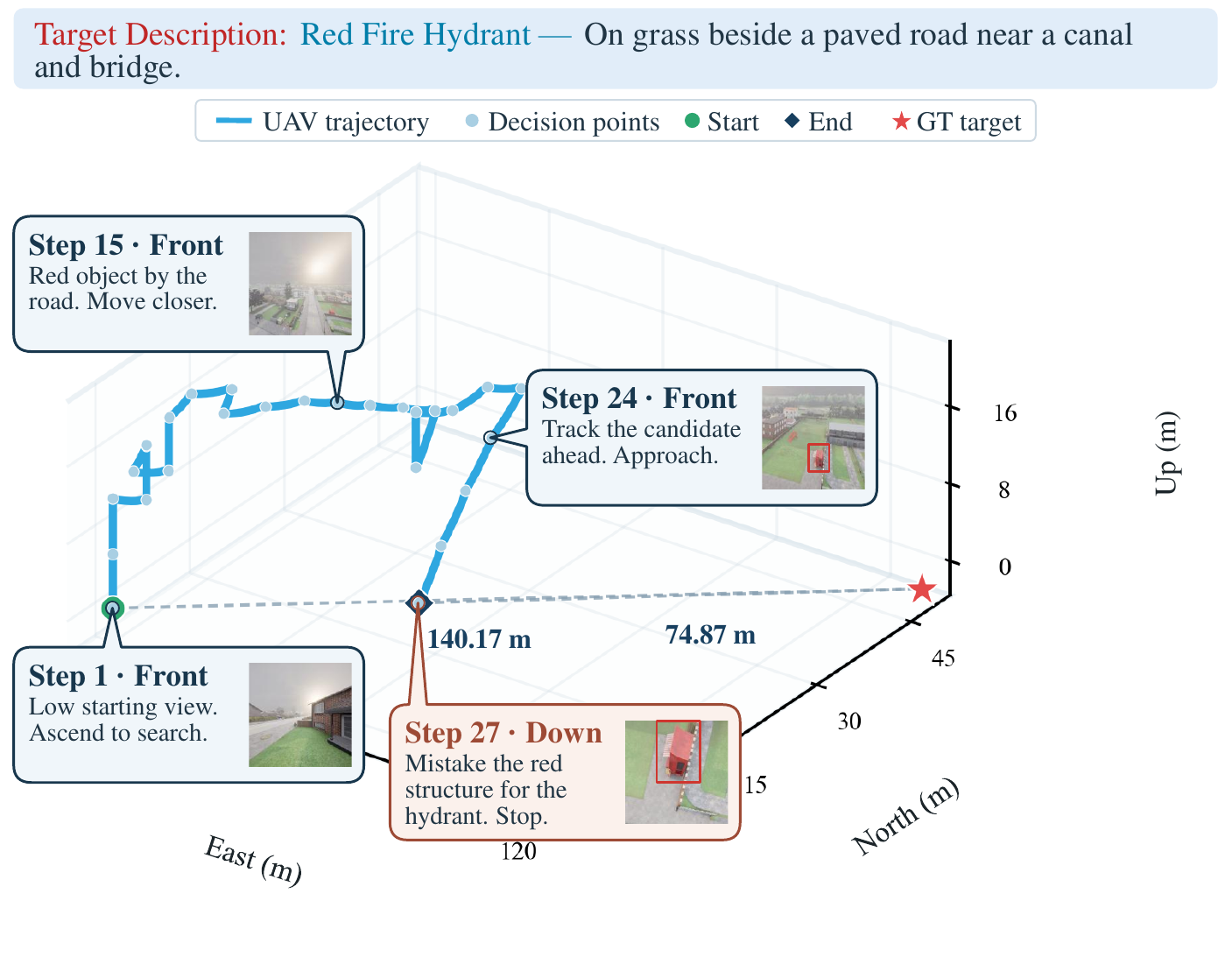}
    \vspace{-2.0em}
    \caption{\textbf{Red fire hydrant search.} A red roadside object prompts an approach, but the UAV stops without clear evidence of the hydrant’s characteristic body or valves, 74.87 m from the annotated target.}
    \label{fig:traveluav_case_hydrant}
  \end{subfigure}
  \vspace{-0.5em}
  \caption{\textbf{Case studies on TravelUAV.} Each panel shows the executed trajectory, selected observations and decisions, and \mbox{annotated target location.}}
  \label{fig:traveluav_case_studies}
\end{figure*}

\subsection{UAV-ON}
\label{app:uavon_case_studies}

\paragraph{Kiosk Search.}
In Figure~\ref{fig:case_study_kiosk}, the UAV first scans the plaza (Step 2) and approaches a red rectangular structure (Step 9). A closer view reveals a telephone booth (Step 13), prompting it to resume searching to the left. At Step 18, the visible open counter and metal sides support identification of the kiosk, and the UAV stops.
This episode illustrates how closer inspection can reject an initially plausible candidate and redirect exploration.

\paragraph{Slide Search.}
In Figure~\ref{fig:case_study_slide}, a possible play area in the right view guides exploration (Step 1). The UAV changes altitude to inspect a possible slide below (Step 10), then ascends after its approach is blocked (Step 15). The resulting downward view reveals the curved metal surface and wooden sides (Step 20), supporting \mbox{confirmation and stopping}.

\paragraph{Person Search.}
In Figure~\ref{fig:case_study_person}, the goal describes a bald person wearing a dark blue jacket, black pants, and black shoes. Two distant figures provide an initial cue, although their clothing is unclear (Step 4). The UAV approaches the figures in the right view (Step 8). A light head and dark top become distinguishable from above (Step 17), motivating closer inspection. At Step 23, the visible bald head and matching clothing support target confirmation and stopping.

\paragraph{Pig Search.}
Figure~\ref{fig:case_study_pig} illustrates an unsuccessful episode. The UAV initially explores the open beach (Step 1), then approaches a pinkish, animal-like object in the downward view (Step 61) and a four-legged statue in the front view (Step 91). Neither inspection yields a confirmed match. The agent remains in search mode when its 150-decision budget expires.

\paragraph{Remarks.}
These examples illustrate how new viewpoints help resolve uncertain candidates and how execution feedback supports recovery after blocked motion.
The failed beach search also shows that sustained exploration and repeated inspection do not guarantee target discovery within the available budget.

\subsection{AerialDojo-200K}
\label{app:aerialdojo_case_studies}

\paragraph{Wooden Chair Search.}
In Figure~\ref{fig:aerialdojo_case_chair}, the UAV detects a possible chair in the front view, but its initial approach is blocked (Step 1). A chair-like shape in the left view prompts a leftward movement (Step 5). Subsequent downward views reveal the shield-shaped back (Step 17), followed by the broken frame and missing seat (Step 18). These attributes match the described damaged chair, and the UAV stops 1.46 m from the target.

\paragraph{Wooden Door Search.}
In Figure~\ref{fig:aerialdojo_case_door}, a downward view reveals a wooden panel lying flat on broken tiles (Step 3). Its appearance and position on the tiles match details in the target description, giving the UAV a candidate for closer inspection. After its first approach is blocked, the UAV repositions to approach the same candidate from the front (Step 8). This route lets the UAV get closer to the panel and confirm it as the door. It stops 1.88~m from the target (Step 10).

\paragraph{Wooden Table Search.}
In Figure~\ref{fig:aerialdojo_case_table}, no table is visible in the front view at Step 2. A possible table later appears on the right, prompting an approach (Step 5). Closer downward views reveal the weathered tabletop (Step 9), followed by its dark surface and straight legs (Step 11). These attributes support confirmation, and the UAV stops 2.42 m from the target.

\paragraph{Brass Candlestick Search.}
Figure~\ref{fig:aerialdojo_case_candlestick} illustrates a failure despite correct target identification. The UAV recognizes the brass candlestick on the mantel, but its direct approach is blocked (Step 4). An erroneous depth estimate places the candidate within the controller’s stopping threshold, leading to a stop at Step~6. The actual distance to the annotated target is 3.87\,m, outside the 3 m success radius.

\paragraph{Remarks.}
The successful episodes illustrate how viewpoint changes and revised approaches support target verification and proximity. The candlestick failure highlights the separate role of geometric accuracy: correct semantic identification can still lead to an unsuccessful stop when the estimated distance is inaccurate.

\subsection{TravelUAV}
\label{app:traveluav_case_studies}

\paragraph{Purple Motorcycle Search.}
In Figure~\ref{fig:traveluav_case_motorcycle}, the bridge and yellow lane markings focus the search on the road segment specified by the instruction (Step 20). Downward views reveal the candidate’s purple color and motorcycle shape as the UAV approaches (Steps 21–23). Starting 78.54 m from the target, it stops at a distance of 3.24 m.

\paragraph{Search for a Man in a Dark Cap.}
In Figure~\ref{fig:traveluav_case_man}, a red tent provides a landmark that guides the UAV along the road (Step 13). The front view then reveals a man wearing a white long-sleeved shirt, gray shorts, and a dark cap (Step 15). These attributes match the instruction, supporting identification and a stop 11.79 m from the target.

\paragraph{Blue Bicycle Search.}
In Figure~\ref{fig:traveluav_case_bicycle}, the UAV starts 354.44 m from the target and explores a wide area. Buildings in the left view prompt it to turn and search nearby (Step 98). After spotting a blue bicycle below, it descends for closer inspection (Step 101). The downward view \mbox{confirms} the bicycle’s color and shape, and the UAV stops 4.37\,m from the target (Step 103).

\paragraph{Red Fire Hydrant Search.}
Figure~\ref{fig:traveluav_case_hydrant} illustrates a false target confirmation. The UAV approaches a red object on grass beside a paved road (Steps 15–24), whose color and location make it a plausible candidate. However, the closer downward view shows a box-shaped object without clear evidence of the upright body or valve outlets characteristic of a hydrant (Step 27). The agent nevertheless stops, 74.87~m from the annotated target.

\paragraph{Remarks.}
These episodes illustrate how landmarks narrow exploration and closer views reveal attributes needed for identification. The hydrant failure shows that matching color and scene context can still mislead the VLM when the target’s characteristic shape remains unverified.

\endgroup

\clearpage
\endgroup

\end{document}